\pdfoutput=1

\documentclass[11pt]{article}

\usepackage[]{ACL2023}
\usepackage{xspace}
\usepackage{times}
\usepackage{latexsym}

\usepackage[T1]{fontenc}

\usepackage[utf8]{inputenc}

\usepackage{microtype}
\usepackage{enumitem}
\usepackage[para]{threeparttable}
\usepackage{booktabs}
\usepackage{multirow}
\usepackage{graphicx}
\usepackage{amsmath}
\usepackage{amssymb}
\usepackage{hyperref}
\usepackage{xcolor}
\usepackage{colortbl}
\usepackage{afterpage}
\usepackage{longtable}
\usepackage{supertabular}
\usepackage{longfbox} 
\usepackage{bm}
\usepackage{makecell}
\usepackage{newtxmath,newtxtext}

\usepackage{etoolbox}

\usepackage{circledsteps}
\definecolor{circlenumbercolor}{HTML}{B83676}

\newcommand\circledBody[2][]{
\hspace{-3pt}\Circled[inner ysep=5pt, fill color=circlenumbercolor, outer color=circlenumbercolor, inner color=white,#1]{\small\textbf{#2}}\hspace{-3pt}
}

\usepackage{inconsolata}

\def\tabmc#1{\multicolumn{1}{c}{#1}}

 \def\tablabel#1{\label{tab:#1}\label{p:#1}}

\def\generalmodelname{Glot500\xspace}
\def\modelname{\mbox{Glot500-m}\xspace}
\def\modelnamejustone{Glot+1\xspace}
\def\corpusnameraw{\mbox{Glot2000-c}\xspace}
\def\corpusnamerawsize{\mbox{700GB}\xspace}
\def\corpusnameclean{\mbox{Glot500-c}\xspace}
\def\corpusnamecleansize{\mbox{600GB}\xspace}

\def\seenlanguage{head language\xspace}
\def\seenlanguages{head languages\xspace}

\def\seenlanguagescripts{head language-scripts\xspace}
\def\seenlanguagesadj{head\xspace}

\def\unseenlanguages{tail languages\xspace}

\def\unseenlanguagescripts{tail language-scripts\xspace}
\def\unseenlanguagesadj{tail\xspace}

\def\numberlanguagesraw{2266\xspace}
\def\numberlanguagescriptclean{534\xspace}

\def\numberlanguagesclean{511\xspace}

\def\languagescript{language-script\xspace}
\def\languagescriptcap{Language-Script\xspace}
\def\languagescripts{language-scripts\xspace}
\def\languagescriptscap{Language-Scripts\xspace}
\def\ngram{n-gram\xspace}
\def\ngramcap{Ngram\xspace}
\def\numberoftasks{six\xspace}

\def\xlmr{XLM-R\xspace}
\def\xlmrb{XLM-R-B\xspace}
\def\xlmrl{XLM-R-L\xspace}

\def\dsource{data source\xspace}
\def\dsources{data sources\xspace}

\def\roundtrip{roundtrip\xspace}
\def\roundtrips{roundtrips\xspace}
\def\Roundtrip{Roundtrip\xspace}

\title{\generalmodelname:\\ Scaling Multilingual Corpora and Language Models to 500 Languages}

\author{Ayyoob Imani$^*$$^{1,2}$, Peiqin Lin$^*$$^{1,2}$, Amir Hossein Kargaran$^{1,2}$, Silvia Severini$^1$, \\ 
  \textbf{Masoud Jalili Sabet$^1$, Nora Kassner$^{1,2}$, Chunlan Ma$^{1,2}$,} \\ \textbf{Helmut Schmid$^1$, André F. T. Martins$^{3,4,5}$, François Yvon$^6$ and Hinrich Schütze$^{1,2}$} \\
        $^1$CIS, LMU Munich, Germany \quad
        $^2$Munich Center for Machine Learning (MCML), Germany \\
        $^3$Instituto Superior Técnico (Lisbon ELLIS Unit)\quad
        $^4$Instituto de Telecomunicações \\
        $^5$Unbabel \quad 
        $^6$Sorbonne Université, CNRS, ISIR, France \\
        \texttt{\{ayyoob, linpq, amir, silvia\}@cis.lmu.de}
}

\def\figref#1{Figure~\ref{fig:#1}}

\def\tablabel#1{\label{tab:#1}\label{p:#1}}

\def\secref#1{\S\ref{sec:#1}}
\def\seclabel#1{\label{sec:#1}}
\def\qref#1{Eq.~\ref{eqn:#1}}

\def\eqlabel#1{\label{eqn:#1}}

\begin{document}
\maketitle
\def\thefootnote{*}\footnotetext{Equal contribution.}\def\thefootnote{\arabic{footnote}}
\begin{abstract}
The NLP community has mainly focused on scaling Large
Language Models (LLMs) \emph{vertically}, i.e.,
making them better for about 100 languages.
We  instead scale
LLMs \emph{horizontally}: we create, through continued pretraining, \modelname, an LLM that covers
\numberlanguagesclean predominantly low-resource languages.
An important part of this effort is to
collect and clean
\corpusnameclean, a corpus that covers these
\numberlanguagesclean
languages and allows us to train \modelname.
We evaluate \modelname
on five diverse tasks across 
these languages.
We observe large improvements for
both high-resource and low-resource languages
compared to an XLM-R baseline.
Our analysis shows that no single
factor explains the quality of multilingual
LLM representations.
Rather, a combination of factors determines quality including
corpus size, script, ``help'' from related languages and the
total capacity of the model.
Our work addresses an important goal of NLP research: we
should
not limit NLP to a small fraction of the world's languages and
instead strive to support as many
languages as possible to bring the benefits of NLP
technology to all languages and cultures.
Code, data and models are available at \url{https://github.com/cisnlp/Glot500}.
\end{abstract}

\section{Introduction}
The NLP community has mainly focused on scaling Large
Language Models (LLMs) \emph{vertically}, i.e.,
deepening their understanding of high-resource languages by
scaling up parameters and training data.  
While this approach has revolutionized NLP,
the achievements are largely limited to high-resource languages.
Examples of ``vertical'' LLMs are GPT3 \cite{brown2020language},
PaLM \cite{chowdhery2022palm} and 
Bloom \cite{lescao2022bloom}. 
In this paper, we
create \modelname, a model that
instead focuses on scaling multilingual
LLMs \emph{horizontally}, i.e., 
scaling to a large number of languages the great majority of
which is low-resource.
As LLMs are essential for progress in NLP,
lack of LLMs supporting low-resource languages
is a serious impediment to bringing NLP to all of the
world's languages and cultures. Our goal is to address this
need with the creation of \modelname.\footnote{In
concurrent work, \citet{adebara2022serengeti}  train a
multilingual model  for 517 African languages on a 42
gigabyte corpus, but without making the model available.}

Existing multilingual LLMs support only about  
100  \citep{conneau-etal-2020-unsupervised}  out of the 7000  languages of
the world. 
These supported languages are the ones for which large amounts of training data 
are available through projects such as Oscar \cite{suarez2019asynchronous}  and 
the Wikipedia dumps.\footnote{\url{https://dumps.wikimedia.org/}}
Following \citet{siddhant2022towards}, we refer to
the 100 languages covered by XLM-R
\citep{conneau-etal-2020-unsupervised}
as \textbf{\seenlanguages} and to
the
remaining languages as \textbf{\unseenlanguages}. 
This terminology is motivated by
the skewed distribution of available data per language: for the
best-resourced languages there are huge corpora available,
but for the long tail of languages, only small corpora exist.
This is a key problem we address:
the availability of data
for \unseenlanguages
is limited compared to \seenlanguages.
As a result, \unseenlanguages have often been ignored by language technologies \cite{joshi-etal-2020-state}.

Although there exists some work on machine translation for a
large number of \unseenlanguages
\citep{costa2022no, bapna2022building},
existing LLMs for \unseenlanguages are limited to a
relatively small number of languages
\cite{wang-etal-2019-improving, alabi-etal-2022-adapting, wang-etal-2022-expanding}.
In this paper, we address this gap.
Our work has three parts. (i) \textbf{Corpus collection.} We
collect \corpusnameraw,
a corpus covering thousands of \unseenlanguages.
(ii) \textbf{Model training.} Using \corpusnameclean, a
subset of \corpusnameraw,
we train \modelname, an
LLM covering \numberlanguagesclean languages.
(iii) \textbf{Validation.} We conduct an extensive evaluation of the quality of \modelname's
representations of \unseenlanguages on 
a diverse suite of tasks.

In more detail, \textbf{corpus collection} considers
three major sources: 
websites that are known to publish content in specific languages, 
corpora with classified multilingual content
and datasets published in specific \unseenlanguages. 
The resulting dataset \corpusnameraw
comprises \corpusnamerawsize  in \numberlanguagesraw languages 
collected from $\approx$150 sources. After cleaning and deduplication, 
we create the subset \corpusnameclean, consisting
of
\numberlanguagesclean languages and 
\numberlanguagescriptclean \emph{language-scripts} (where  we define
a language-script as a combination of ISO 639-3\footnote{\url{https://iso639-3.sil.org/code_tables/639}} and script)
to train \modelname.
Our criterion for including a \languagescript
in \corpusnameclean is that it includes 
more than 30,000 sentences.

\textbf{Model training.}
To train \modelname, we employ vocabulary extension and
continued pretraining. XLM-R's vocabulary is extended with
new tokens trained on \corpusnameclean. We then perform continued pretraining of XLM-R with the MLM objective \citep{devlin-etal-2019-bert}.

\textbf{Validation.}
We comprehensively evaluate \modelname on
a diverse suite of natural language understanding, sequence
labeling and multilingual
tasks for 
hundreds of languages. The results demonstrate that \modelname
performs better than \xlmrb (XLM-R-base) for \unseenlanguages
by a large margin while performing comparably (or better) for \seenlanguages.

Previous work on multilinguality has been hindered by the
lack of LLMs supporting a large number of languages.  This
limitation has led to studies being conducted in settings
dissimilar from real-world scenarios.  For
example, \citet{dufter-schutze-2020-identifying} use
synthetic language data. And the curse of multilinguality
has been primarily studied for a set of high-resource
languages \citep{conneau-etal-2020-unsupervised}.  By
creating \modelname, we can investigate these issues in a
more realistic setting. We make code, data and trained
models available to foster  research by
the community on how to include hundreds of
languages that are currently ill-served by NLP technology.

\textbf{Contributions.}
(i) We train the multilingual model \modelname on a
\corpusnamecleansize 
corpus, covering more than 500 diverse languages,
and make it publicly available at \url{https://github.com/cisnlp/Glot500}.
(ii) We
collect and clean
\corpusnameclean, a corpus that covers these diverse languages
and allows us to train \modelname, and will make as much of it
publicly available as possible.
(iii)
We evaluate \modelname
on pseudoperplexity and on five diverse tasks across 
these languages.
We observe large improvements for
low-resource languages
compared to an XLM-R baseline.
(iv)
Our extensive analysis shows that no single
factor explains the quality of multilingual
LLM representations.
Rather, a combination of factors determines quality including
corpus size, script, ``help'' from related languages and the
total capacity of the model.
(v)
Our work addresses an important goal of NLP research: we
should
not limit NLP to a relatively small number of high-resource languages and
instead strive to support as many
languages as possible to bring the benefits of NLP
to all languages and cultures.

\section{Related Work}
Training
multilingual LLMs using the masked
language modeling (MLM) objective is effective
to achieve cross-lingual representations
\citep{devlin-etal-2019-bert, conneau-etal-2020-unsupervised}.
These models can be 
further improved by incorporating techniques such as discriminative pre-training
\citep{chi-etal-2022-xlm} and the use of parallel data \citep{yang2020alternating, chi-etal-2021-infoxlm}. 
However, this primarily benefits a limited set of 
languages with large corpora.

Recent research has attempted  to extend existing LLMs to 
languages with limited resources. \citet{wang-etal-2019-improving} 
propose vocabulary extension; \citet{ebrahimi-kann-2021-adapt} investigate 
adaptation methods, including MLM and 
Translation Language Model (TLM) objectives and adapters;
\citet{alabi-etal-2022-adapting} adapt XLM-R to 
17 African languages;
\citet{wang-etal-2022-expanding} expand language models to 
low-resource languages using bilingual lexicons.

Alternatively,
parameter-efficient fine-tuning adapts pre-trained models to new languages 
by training a small set of weights effectively 
\citep{zhao-etal-2020-masking,pfeiffer-etal-2021-unks,ansell-etal-2022-composable}. 
\citet{pfeiffer-etal-2022-lifting} address the ``curse of multilinguality''
by sharing a part of the model among all languages and 
having separate modules for each language.
We show  that 
the common 
perception that multilinguality increases as we add more
languages, until, from some point,
it starts decreasing,
is naive. The amount of available data per language and the similarity between
languages also play important roles (\secref{supportthroughrelatedlanguages}).

Another approach trains LLMs from scratch for a
limited number of 
\unseenlanguages; e.g.,
AfriBERTa \citep{ogueji-etal-2021-small}
and IndicNLPSuite \citep{kakwani-etal-2020-indicnlpsuite}
are LLMs
for 11 African 
languages and 11 Indic languages.
In concurrent work, \citet{adebara2022serengeti}  train a
multilingual model  for 517 African languages on a 42
GB corpus, but without making the model available and
with an evaluation on a smaller number of languages than ours.

Closely related to our work on corpus creation,
\citet{bapna2022building} and \citet{costa2022no}
also create NLP resources for a large number of \unseenlanguages.
They
train a language identifier model and extract textual data for \unseenlanguages
from large-scale web crawls. This approach is effective, 
but it requires significant computational resources and native speakers for 
all \unseenlanguages. This is hard to do outside of large corporations. 
\citet{bapna2022building} have not made their data available.
\citet{costa2022no} have only 
released a  portion of their data in around 200 languages.

A key benefit of ``horizontally'' scaled multilingual LLMs
is transfer from high- to low-resource
languages. Our evaluation suggests that \modelname excels at
this, but this is not the main focus of our paper. There is
a large body of work on crosslingual transfer:
\citep{artetxe-schwenk-2019-massively,imanigooghari-etal-2022-graph, lauscher-etal-2020-zero,conneau-etal-2020-unsupervised,DBLP:journals/corr/abs-2106-16171,
DBLP:journals/jmlr/FanBSMEGBCWCGBL21, severini2022towards,
choenni-shutova-2022-investigating,DBLP:journals/corr/abs-2305-00090},
inter alia.

\section{\corpusnameraw}

\subsection{Data Collection}
One of the major challenges in developing NLP technologies 
for \unseenlanguages is the scarcity of high-quality training data. 
In this work, we propose a lightweight methodology that is
easily replicable for academic labs.
We identify tail language data previously published by researchers, publishers
and translators and then crawl or download them.
By crawling a few websites and 
compiling data from around 150 different datasets, we
amass more than \corpusnamerawsize  of 
text in \numberlanguagesraw 
languages.
We will refer to these sources of data
as \emph{\dsources}. 
Our data covers many domains, including religious texts,
news articles and scientific 
papers.
Some of the \dsources are high-quality, verified 
by native speakers, translators and linguists. Others are less reliable 
such as web crawls and Wikipedia dumps. 
It is therefore necessary to clean
the data.
For a list of \dsources, see \secref{appendix_datasets}.

\subsection{\languagescriptscap}
Some languages are written in multiple scripts;
e.g., Tajik  is written in both Cyrillic and Arabic scripts. 
Some \dsources indicate the script, 
but others either do not  or provide mixed text
in multiple scripts.
We detect the script for
each sentence 
and treat each \languagescript as a separate entity.

\subsection{\ngramcap LMs and Language Divergence}
\seclabel{ngramlms}
We train a 3-gram character-level language model $M_i$ for each language-script $L_i$,
using KenLM \cite{heafield-2011-kenlm}.
We refer to the perplexity calculated 
for the corpus of language $L_i$ using language model $M_j$
as $\mathcal{PP}(M_j, L_i)$.
Similar to \citet{gamallo-etal-2017-perplexity}, 
we define a perplexity-based 
divergence measure
of languages $L_i$ and $L_j$ as:
\begin{equation*}
  \resizebox{0.89\hsize}{!}{$\mathcal{D}_{L_i, L_j}= 
  \max \big(\mathcal{PP}(M_j, L_i), \mathcal{PP}(M_i, L_j) \big)$}
\end{equation*}
We use $\mathcal{D}$ to filter out noisy
data in \secref{datacleaning} and
study
the effect of similar languages in  LLM training
in   \secref{AmountoftrainingDataeffect} and 
\secref{supportthroughrelatedlanguages}.
For more details,
see \secref{appendix_ngrams}.

\subsection{Data Cleaning}
\seclabel{datacleaning}
To remove noise, we use
chunk-level and
corpus-level filters.

While some sources  are
sentence-split,
others provide multiple sentences (e.g., a paragraph)
as one chunk. 
Chunk-level filters process each chunk of text 
from a \dsource  as a unit, without sentence-splitting.
Some chunk-level filters are based on the notion of word: we use
white space tokenization when possible and otherwise resort to
sentencePiece \citep{kudo-richardson-2018-sentencepiece}
trained by \citet{costa2022no}.

As chunk-level filters, we employ the \textbf{sentence-level
filters} SF1--SF5 from BigScience ROOTS
\citep{laurencconbigscience}.

  \textbf{SF1} Character repetition. If the ratio of repeated characters  is too high, 
    it is likely that the sentence has not enough textual content.

\textbf{SF2} Word repetition. A high ratio of repeated words indicates non-useful repetitive content. 

\textbf{SF3} Special characters. Sentences with a high ratio of special 
    characters are likely to be crawling artifacts or computer code.

\textbf{SF4} Insufficient number of words. Since training language models requires enough context, very small chunks of text are not useful.

\textbf{SF5} Deduplication. 
    If two sentences are identical
    after eliminating punctuation and white space,
    one is removed.

In the rest of the paper, we refer to a chunk as
a \textbf{sentence'}. A sentence' can consist of a short segment, a
complete sentence or a chunk (i.e., several sentences).

\textbf{Corpus-level
filters} detect if  
the corpus of a \languagescript is noisy;
e.g., the corpus is 
in another language or 
consists of non-meaningful  content such as tabular data.
We employ filters CF1 and CF2.

\textbf{CF1}
In case of \textbf{mismatch between language and script}, 
the corpus is removed; e.g., Chinese 
    written in Arabic  is unlikely to be Chinese.

  \textbf{CF2}\label{LF:Perplexity} Perplexity mismatch.
For each \languagescript L1, we find its closest
\languagescript L2: the
\languagescript with the lowest
perplexity divergence (\secref{ngramlms}).
If L1 and L2 are not in the
same typological family, we
check L1/L2 manually and take appropriate action such as
removing the corpus (e.g., if it is actually English)
or correcting the ISO code assigned to the corpus.

\def\csimangle{75}
\begin{table}[t]
  \small
	\centering
	\def\tablesep{0.1cm}
\begin{tabular}{
  @{\hspace{\tablesep}}l@{\hspace{\tablesep}}|
  @{\hspace{\tablesep}}r@{\hspace{\tablesep}}
  @{\hspace{\tablesep}}r@{\hspace{\tablesep}}
  @{\hspace{\tablesep}}r@{\hspace{\tablesep}}
  @{\hspace{\tablesep}}r@{\hspace{\tablesep}}
}
  & \rotatebox{\csimangle}{langs}
  & \rotatebox{\csimangle}{scripts}
  & \rotatebox{\csimangle}{sents'}
  & \rotatebox{\csimangle}{median s'} \\
  \midrule
\corpusnameraw & \numberlanguagesraw & 35 & 2.3B & 8K \\
\corpusnameclean & \numberlanguagesclean & 30 & 1.5B & 120K \\
  \citet{costa2022no} & 134 & - & 2.4B & 3.3M \\
  \citet{bapna2022building} & 1503 & - & 1.7B & 25K 
\end{tabular}
\caption{Statistics
for \corpusnameraw,
\corpusnameclean and existing 
  multilingual datasets: number of languages,  scripts,
  sentences' and median number of sentences' per language-script.
}
\tablabel{data_stats}
\end{table}

\subsection{Training Data: \corpusnameclean}
Among the 2000+ language-scripts that we collected data for, 
after cleaning, most
have too little data for
pretraining LLMs. It is difficult
to quantify the minimum amount needed for
pretraining.  Therefore, we pick a relatively high ``safe'' threshold,
30,000 sentences', for inclusion of \languagescripts in model training.
This allows us to train the model effectively and cover
many low-resource languages.
Table~\ref{tab:data_stats} gives
\corpusnameclean statistics.
See \secref{appendix_languages} 
for a  list of \languagescripts.
We train   \modelname
on \corpusnameclean; note that while \corpusnameclean 
focuses on \unseenlanguages, it contains some data in \seenlanguages which 
we include in \modelname training to prevent catastrophic forgetting.

We divide the corpus for each language into train/dev/test,
reserving 1000 sentences' each for dev and test
and using the rest for train.
We pick 1000 parallel  verses 
if we have a Bible translation
and add 500 each to test and dev.
These parallel verses convey identical meanings and
facilitate crosslingual evaluation.
We pretrain the model using only the training data.

\section{\modelname}
\subsection{Vocabulary Extension}
\seclabel{vocabulary_extension}
To extend XLM-R's vocabulary,
we use SentencePiece \cite{kudo-richardson-2018-sentencepiece} with a unigram
language model \cite{kudo-2018-subword} to train a tokenizer with a
vocabulary size of 250K on \corpusnameclean. We sample data from
different \languagescripts according to a multinomial distribution,
with $\alpha$=.3. 
The amount we sample for \seenlanguages is the
same as \unseenlanguages with the lowest amount; this
favors  \unseenlanguages\ -- \seenlanguages are already
well learned by XLM-R.
We merge the obtained tokens with XLM-R's vocabulary.
About 100K new tokens
were in fact old
tokens, i.e.,  already part of XLM-R's vocabulary.
We take the
probabilities of the (genuinely) new tokens directly from
SentencePiece.
After adding the 151K  new tokens to XLM-R's vocabulary
(which has size 250K), the vocabulary
size of \modelname is 401K.

We could also
calculate probabilities of  existing and new tokens over
a mixture of original \xlmr training corpus and \corpusnameclean 
\citep{chung-etal-2020-improving}. 
For \seenlanguages, the percentage of changed tokens using 
the new tokenizer compared to the original tokenizer ranges from 0.2\%
to 50\%. However,  we found no relationship between percentage of changed
tokens and change in  performance on downstream tasks.
Thus, there was little effect of tokenization in our experiments.

\begin{table}[t]
  \small
	\centering
	\def\tablesep{0.1cm}
\begin{tabular}{
  @{\hspace{\tablesep}}l@{\hspace{\tablesep}}|
  @{\hspace{\tablesep}}r@{\hspace{\tablesep}}
  @{\hspace{\tablesep}}r@{\hspace{\tablesep}}
  @{\hspace{\tablesep}}c@{\hspace{\tablesep}}
}
                            & \xlmrb & \xlmrl & \modelname \\
  \midrule
Model Size    & 278M & 560M  & 395M \\
Vocab Size      & 250K & 250K & 401K \\
Transformer Size      & 86M & 303M & 86M 
  \end{tabular}
  \caption{Model sizes. \modelname and  \xlmrb have the
  same transformer size, but \modelname has a larger
  vocabulary, resulting in an overall  larger model.}
  \tablabel{parameter}
\end{table}

\subsection{Continued Pretraining}
\seclabel{continued}
We create \modelname by continued pretraining
of \xlmrb 
with the MLM objective. 
The optimizer used is Adam with betas (0.9, 0.999).
Initial learning rate: 5e-5.
Each training step contains a batch of 384 training samples randomly picked from all \languagescripts.
The sampling strategy across \languagescripts is the same as
for vocabulary extension (\secref{vocabulary_extension}).
We save checkpoints every 10K steps and select the
checkpoint with the best average performance on downstream tasks by
early stopping. Table \ref{tab:parameter} lists the 
sizes of \xlmrb, \xlmrl and \modelname.
Except for a larger vocabulary (\secref{vocabulary_extension}),
\modelname 
has the same size
as \xlmrb. We train \modelname on a server with 
eight NVIDIA RTX A6000 GPUs for two weeks.

Similar to XLM-R, we concatenate sentences' of a \languagescript and
feed them as a stream to the
tokenizer. The resulting output is then divided into chunks
of 512 tokens and fed to the model.

\begin{table}[t]
  \small
	\centering
	\def\tablesep{0.1cm}
\begin{tabular}{
  @{\hspace{\tablesep}}l@{\hspace{\tablesep}}|
  @{\hspace{\tablesep}}r@{\hspace{\tablesep}}
  @{\hspace{\tablesep}}r@{\hspace{\tablesep}}
  @{\hspace{\tablesep}}c@{\hspace{\tablesep}}
}
                            & |head| & |tail| & measure (\%) \\
  \midrule
Sentence Retrieval Tatoeba    & 70 & 28  & Top10 Acc. \\
Sentence Retrieval Bible      & 94 & 275 & Top10 Acc. \\
  Text Classification        & 90 & 264 & F1 \\
  NER                        & 89 & 75  & F1 \\
  POS                        & 63 & 28  & F1 \\
  \Roundtrip Alignment     & 85 & 288 & Accuracy
  \end{tabular}
  \caption{Evaluation tasks and measures.
|head|/|tail|: number of head/tail 
  \languagescripts}
  \tablabel{evaluationmetrics}
\end{table}

\section{Experimental Setup}
For most \unseenlanguages, there are no manually labeled evaluation data.
We therefore adopt a mixed evaluation
strategy: based partly on human labels, partly on
evaluation methods 
that are applicable to many languages 
without requiring gold data. 
Table~\ref{tab:evaluationmetrics} lists all our evaluation 
tasks.

\paragraph{Perplexity}
Following \citet{salazar-etal-2020-masked}, we calculate pseudoperplexity (PPPL) 
 over the held-out test set.
PPPL is based on masking tokens one-by-one (not left to right).
\citet{salazar-etal-2020-masked} give evidence that PPPL
is a better measure of linguistic acceptability compared to 
standard left-to-right perplexity.

\paragraph{\Roundtrip Alignment}
For assessing the quality of multilingual 
representations for a broad range of \unseenlanguages
without human gold data, we adopt \roundtrip evaluation
\citep{dufter-etal-2018-embedding}.  We
first word-align sentences' in a parallel corpus based on
the multilingual representations of an LLM.
We then start
from a word $w$ in a sentence' in \languagescript L1, follow the alignment
links to its translations in \languagescript L2,
then the alignment links from L2 to L3 and so on, until in
the end we follow alignment links back to L1.
If this ``\roundtrip'' gets us back to $w$, 
then it indicates that the
LLM has similar representations
for the meaning of $w$ in \languagescripts L1, L2, L3, etc. In
other words, the cross-lingual quality of representations is
high. Vice versa, failure to get back to $w$ is a sign
of poor multilingual representations.

We use SimAlign 
\citep{jalili-sabet-etal-2020-simalign} and align
on the sub-word level on the Bible part of test,
based on the representations of the LLM computed by transformer layer~8 as suggested in the original paper. 
We use intersection 
symmetrization: each word in a sentence' is aligned to 
at most one word in the other sentence'.

As evaluation measure we compute the
percentage of \roundtrips that were successes, i.e., the
\roundtrip starts at $w$ in L1 and
returns back to $w$.
For each \languagescript in test, 
we randomly select three  \languagescripts
as intermediate points L2, L3, L4.
Since the intermediate points influence
the results, we run the experiment five
times with different
intermediate points
and report the average.
All models are evaluated with the same five sets of three
intermediate \languagescripts.

\paragraph{Sequence Labeling}
We consider two sequence labeling tasks: Named Entity
Recognition (NER) and Part-Of-Speech (POS) tagging.
We use the WikiANN
dataset \cite{pan-etal-2017-cross} for NER and version
v2.11 of  Universal Dependencies (UD)
\cite{10.1162/coli_a_00402} for POS.
Since training
data does not exist for some languages, we
finetune on English
(with early stopping based on dev)
and
evaluate
zero-shot transfer on all languages covered by WikiANN/UD.
We set the learning rate to
2e-5 with Adam.

\paragraph{Sentence Retrieval}\label{sec:tasksentenceretrieval}
Following \cite{pmlr-v119-hu20b}, we use up to 1000
English-aligned sentences' from Tatoeba
\cite{artetxe-schwenk-2019-massively}
to evaluate SentRetr
(sentence retrieval). We also use 500 English-aligned sentences' 
from the Bible part of test.
We find nearest neighbors using cosine similarity based on the average word embeddings
in layer $l=8$
-- following \citet{jalili-sabet-etal-2020-simalign} --
and compute
top10 accuracy. For fair comparison
and because the architectures are the same,
we do not optimize the hyperparameter $l$ 
for \modelname and \xlmrb.

\paragraph{Text Classification}
We evaluate on Taxi1500
\citep{ma2023taxi1500}.  It provides gold data for
text classification with six classes in a large number of
\languagescripts of which
\modelname supports
354.
We finetune  on English
(with early stopping on dev)
and
evaluate zero-shot  on test of the target
\languagescript. Learning rate:  2e-5,
batch size: 16 (following \citet{ma2023taxi1500}).

\begin{table*}[t]
  \centering
  \footnotesize
  \def\tablesep{0.05cm}
  \resizebox{\textwidth}{!}{
  \begin{tabular}{
    @{\hspace{\tablesep}}l@{\hspace{\tablesep}}|
    @{\hspace{\tablesep}}r@{\hspace{\tablesep}}
    @{\hspace{\tablesep}}r@{\hspace{\tablesep}}
    @{\hspace{\tablesep}}r@{\hspace{\tablesep}}|
    @{\hspace{\tablesep}}r@{\hspace{\tablesep}}
    @{\hspace{\tablesep}}r@{\hspace{\tablesep}}
    @{\hspace{\tablesep}}r@{\hspace{\tablesep}}|
    @{\hspace{\tablesep}}r@{\hspace{\tablesep}}
    @{\hspace{\tablesep}}r@{\hspace{\tablesep}}
    @{\hspace{\tablesep}}r@{\hspace{\tablesep}}
  }
    & \multicolumn{3}{c}{\unseenlanguagesadj} & \multicolumn{3}{c}{\seenlanguagesadj} & \multicolumn{3}{c}{all} \\
    &  {\xlmrb} & {\xlmrl} & {\modelname} & {\xlmrb} & {\xlmrl} & {\modelname} & {\xlmrb} & {\xlmrl} & {\modelname} \\
    \midrule         
    Pseudoperplexity                 & 304.2 & 168.6 & \bfseries12.2  & 12.5  & \bfseries8.4   &           11.8 & 247.8  & 136.4 & \bfseries 11.6 \\
    Sentence Retrieval Tatoeba \, & 32.6 & 33.6 & \bfseries59.8 & 66.2 & 71.1          & \bfseries75.0 & 56.6  & 60.4 & \bfseries70.7 \\
    Sentence Retrieval Bible   & 7.4 & 7.1 & \bfseries43.2 & 54.2 & 58.3          & \bfseries59.0 & 19.3  & 20.1 & \bfseries47.3 \\
    Text Classification        & 13.7 & 13.9 & \bfseries46.6 & 51.3 & \bfseries60.5 & 54.7          & 23.3  & 25.8 & \bfseries48.7 \\
    NER                        & 47.5 & 51.8 & \bfseries60.7 & 61.8 & \bfseries66.0 & 63.9          & 55.3  & 59.5 & \bfseries62.4 \\
    POS                        & 41.7 & 43.5 & \bfseries62.3 & 76.4 & \bfseries78.4 & 76.0          & 65.8  & 67.7 & \bfseries71.8 \\
    \Roundtrip Alignment       & 2.6  & 3.1 & \bfseries4.5  & 3.4  & 4.1           & \bfseries5.5  & 2.8   & 3.3  & \bfseries4.7            
    \end{tabular}
    }
    \caption{Evaluation of XLM-R base and large (\xlmrb and
  \xlmrl) and \modelname on pseudoperplexity and \numberoftasks multilingual tasks across 5 seeds. Each
  number is an average over 
\seenlanguagesadj, \unseenlanguagesadj and all
  \languagescripts. See \secref{appendix_results},
  \secref{appendix_perplexity_results} for 
  results per task and \languagescript.  
    \modelname outperforms \xlmrb  in all tasks for
    \seenlanguagesadj (except for POS) and \unseenlanguagescripts
and \xlmrl for \unseenlanguagescripts.
Best result per row/column group in bold.}
\tablabel{mainresults}
\end{table*}

\section{Experiments}
In this  section, we discuss aggregate  results.
For detailed results, see \secref{appendix_results} and \secref{appendix_perplexity_results}.
\subsection{Results}
Table~\ref{tab:mainresults} gives results. 
\modelname outperforms \xlmrb
on all tasks for both \seenlanguagesadj
and \unseenlanguagescripts, except for POS on \seenlanguagesadj.
That \modelname
outperforms \xlmrb is
expected for \unseenlanguagescripts (i.e., those not covered by
XLM-R).
For these \languagescripts the improvement margin is large.
Outperformance may seem counterintuitive for \seenlanguagescripts
(those covered by XLM-R)
since \modelname
has the same number of (non-embedding) parameters
as  \xlmrb.
Since the
number of covered languages has greatly increased, leaving
less capacity per language, we might expect underperformance.
There are a few possible explanations.
First, XLM-R may be undertrained, and the inclusion of
more \seenlanguagesadj language 
training data may improve their representations.
Second, having more languages may improve multilinguality by allowing languages 
to synergize and enhance each other's representations and cross-lingual transfer.
Third, there are languages similar to \seenlanguages among the \unseenlanguages,
which in turn aids  \seenlanguages.

The gap between \modelname and the baselines 
for \unseenlanguagescripts in sequence labeling  is smaller. 
These tasks do not require as
deep an understanding of language and thus transfer from \seenlanguagesadj
to \unseenlanguagescripts is easier through shared tokens. 

\modelname also outperforms \xlmrl for \unseenlanguagescripts
(all tasks) and \seenlanguagescripts (3 tasks). 
This suggests that scaling up size is not the only way for 
improvements. We can also improve the quality of
multilingual LLM representations by
increasing the number of languages.

\begin{figure}
  \centering
  \includegraphics[width=0.49\columnwidth]{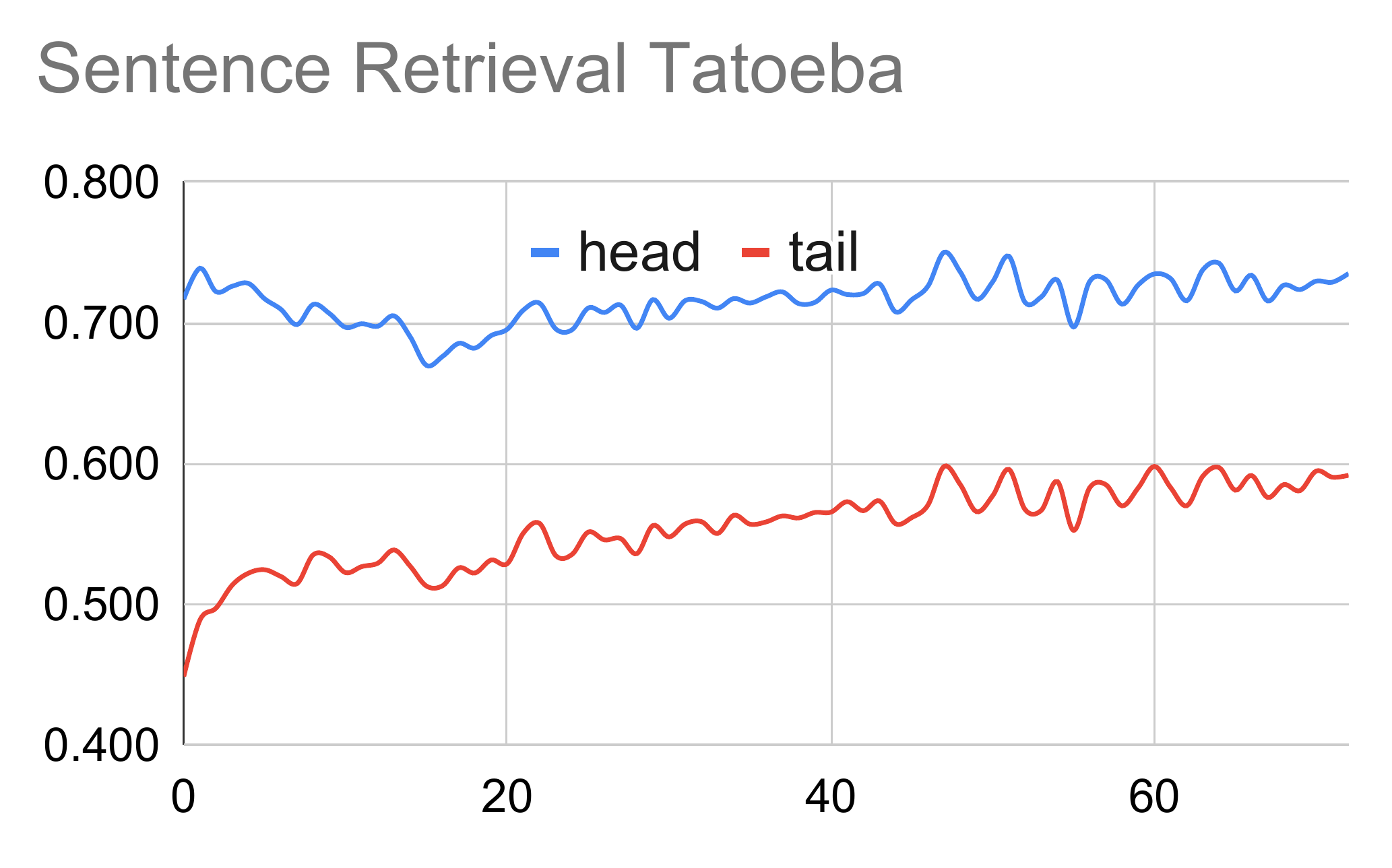}
  \includegraphics[width=0.49\columnwidth]{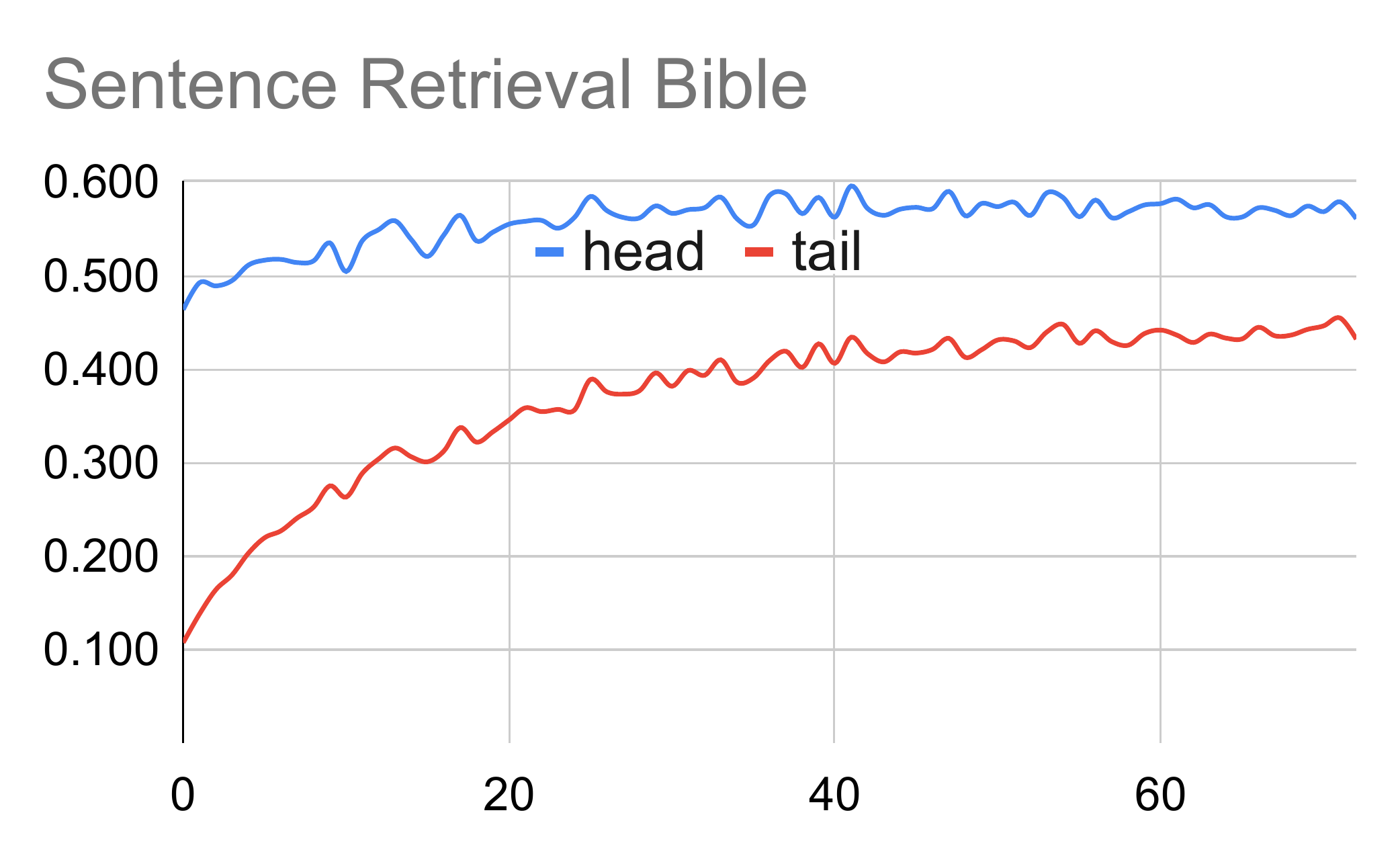}
  \includegraphics[width=0.49\columnwidth]{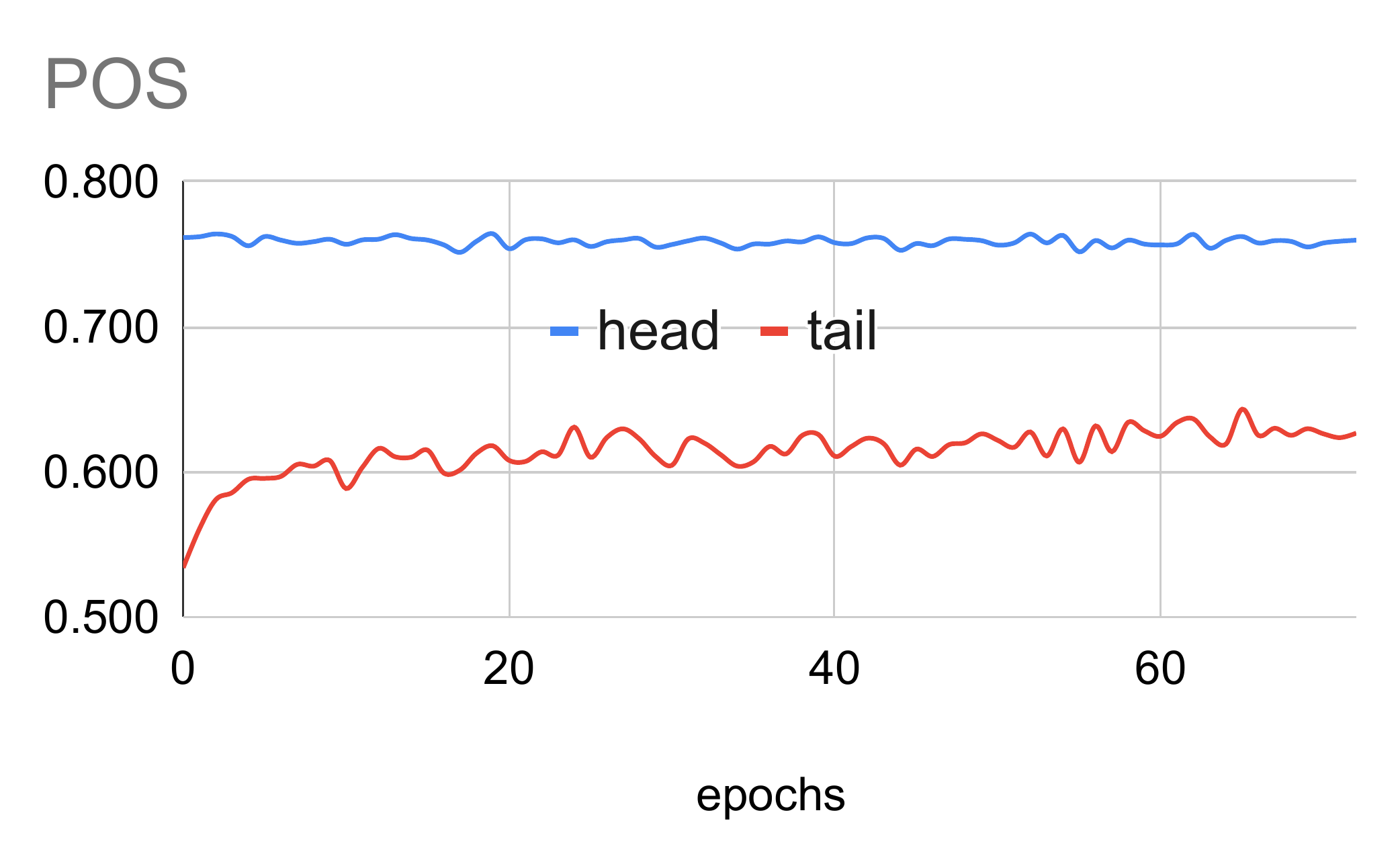}
  \includegraphics[width=0.49\columnwidth]{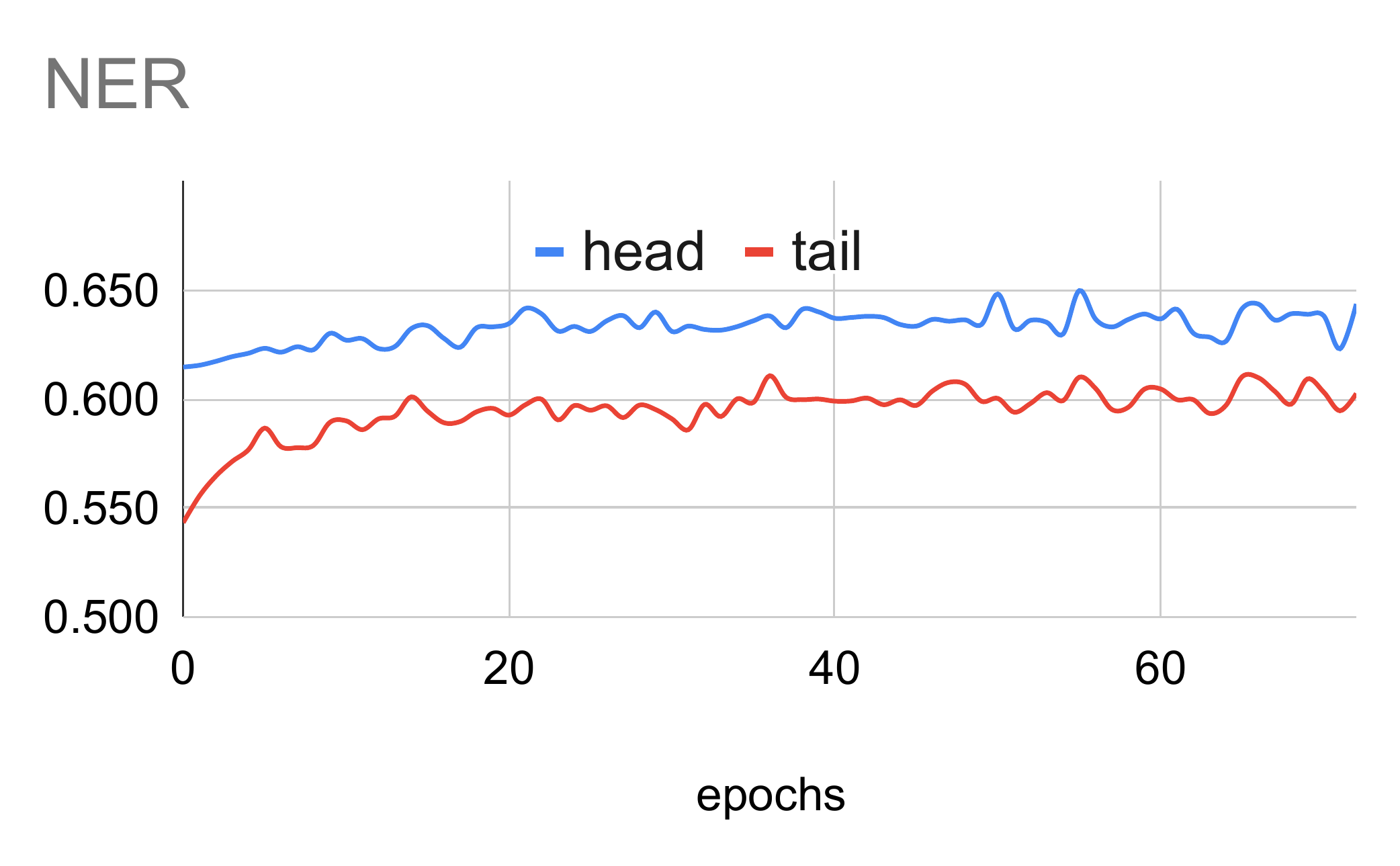}
  \caption{Progression of training for sentence 
  retrieval and sequence labeling.
x-axis: epochs/10K.
The improvement is fast in the beginning for
\unseenlanguages, then gets slower and
and reaches a plateau.
This pattern is partially observed for  \seenlanguages.
  }
  \label{fig:training_progression_unseen}
\end{figure}

\begin{table}[t]
  \centering
  \small
  \begin{tabular}{l|rr}
    & \seenlanguagesadj & \unseenlanguagesadj \\
    \hline
    \modelname is better & 37 & 420 \\
    \xlmrb is better & 69 & 8 
  \end{tabular}
  \caption{Pseudoperplexity \modelname vs \xlmrb. \modelname's worse
  performance on \seenlanguagesadj can be attributed to smaller
  training corpora and the relative difficulty of learning
  five times more languages with the same number of
  (non-embedding) parameters. \modelname performs better on
  almost all \unseenlanguagescripts. \protect\secref{coverage}
  discusses  the eight exceptions.
  \tablabel{perplexitywin}}
\end{table}

\subsection{Language Coverage}
\seclabel{coverage}
Table \ref{tab:perplexitywin} compares \modelname vs.\ \xlmrb
on pseudoperplexity. For  fair comparison we use word-level normalization.
 For 69 \seenlanguagescripts, \modelname
underperforms \xlmrb. This is expected as
\modelname's training data is small for these \languagescripts.
\modelname outperforms \xlmrb for
420 \unseenlanguagescripts.

There are eight \unseenlanguagescripts  for which \modelname
performs worse than \xlmrb. Five 
are \unseenlanguages with a 
similar \seenlanguage where the two share a macro-language:
ekk/Standard Estonian (est/Estonian),
aln/Gheg Albanian (sqi/Albanian), nob/Norwegian Bokmal
(nor/Norwegian), hbs/Serbo-Croatian (srp/Serbian),
lvs/Standard Latvian (lav/Latvian).
Since \xlmrb's pretraining corpus is large for the
five \seenlanguages, its performance is good for the close \unseenlanguages.

The other three languages all have a unique script: sat/Santali (Ol Chiki script), div/Dhivehi (Thaana script), iku/Inuktitut (Inuktitut syllabics).
For these languages, \xlmrb's tokenizer returns many UNK tokens since it is not trained on these scripts, resulting in an unreasonably optimistic estimate of pseudoperplexity by our implementation.

\modelname's token-level normalized pseudoperplexity ranges from 1.95 for
lhu/Lahu
to 94.4 for tok/Toki Pona. The average is 13.5, the median 10.6. 
We analyze the five language-scripts with the highest pseudoperplexity:
tok\_Latn, luo\_Latn, acm\_Arab, ach\_Latn, and teo\_Latn.

tok/Toki Pona is a constructed language. According to
Wikipedia:
``Essentially identical concepts can be described by different
words as the choice relies on the speaker's perception and experience.''
This property can result in higher variability and higher perplexity.

acm/Mesopotamian Arabic  contains a large
number of tweets in raw form. This may  result in 
difficult-to-predict tokens in test.

luo/Luo, ach/Acoli and teo/Teso are related Nilotic
languages spoken in Kenya, Tanzania, Uganda and South Sudan.
Their high perplexity could be related
to the fact that they are tonal languages, but the tones are
not orthographically indicated. Another possible explanation
is that the training data is dominated by one subcorpus
(Jehova's Witnesses) whereas the test data are dominated by
PBC. There are orthographic differences between the two,
e.g., ``dong'' (JW) vs.\
``do\ng{}''
(PBC) for Acoli. These
three languages are also spoken over a large area in
countries with different standard languages, which could
increase variability.

Our analysis is not conclusive. We note however that the gap
between
the three languages and the next most difficult languages in
terms of pseudoperplexity is not large. So maybe 
Luo, Acoli and Teso
are
simply (for reasons still to be determined) languages that
have higher perplexity than others.

\subsection{Training Progression}
To analyze the training process,
we evaluate \modelname  on sequence labeling 
and SentRetr  at 10,000-step intervals.
\figref{training_progression_unseen}
shows
that performance  improves rapidly at the onset 
of training, but then the rate of improvement slows down.
This trend is particularly pronounced for \unseenlanguages in SentRetr. 
In comparison, sequence labeling is  relatively straightforward, 
with the baseline (\xlmrb, epoch 0) achieving high performance by correctly transferring prevalent classes such as \emph{verb} and \emph{noun} through shared vocabulary, 
resulting in a smaller improvement of \modelname vs.\ \xlmrb.

For SentRetr, 
we observe larger improvements for
the Bible than for Tatoeba.
This is likely due to the higher proportion of religious data in \corpusnameclean, 
compared to
XLM-R's training data (i.e., CC100).

The average performance on downstream 
tasks peaks at 480K steps.
We have taken a snapshot of \modelname  at this stage and 
 released it.

\def\isoseparator{0.02cm}
\def\fullnameseparator{0.03cm}
\begin{table*}[t]
  \centering
  \small
  \def\tablesep{0.1cm}

    \begin{tabular}{c|c|l@{\hspace{\isoseparator}}l@{\hspace{\fullnameseparator}}l@{\hspace{0.075cm}}r@{\hspace{0.075cm}}rr|c|l@{\hspace{\isoseparator}}l@{\hspace{\fullnameseparator}}l@{\hspace{0.075cm}}r@{\hspace{0.075cm}}rr}
&    & \multicolumn{3}{c}{language-script} & XLMR & Glot500 & gain & &\multicolumn{3}{c}{language-script} & XLMR& Glot500 & gain \\ 
    \midrule
    \midrule
    \multirow{5}{*}{ \makecell{\rotatebox{90}{high end}}}
 &   \multirow{10}{*}{ \makecell{\rotatebox{90}{SentRetr
    Tatoeba} }} & tat&C& Tatar & 10.3 & 70.3 & 60.0
    & \multirow{10}{*}{ \makecell{\rotatebox{90}{SentRetr
    Bible} }} & uzn&C& Northern Uzbek& 5.4 & 87.0 & 81.6 \\
 &   & nds&L &Low German & 28.8 & 77.1 & 48.3 & & crs&L& Seselwa Creole& 7.4 & 80.6 & 73.2 \\
  &  & tuk&L &Turkmen & 16.3 & 63.5 & 47.3 & & srn&L &Sranan Tongo& 6.8 & 79.8 & 73.0 \\
   & & ile&L &Interlingue & 34.6 & 75.6 & 41.0 & & uzb&C& Uzbek & 6.2 & 78.8 & 72.6 \\
   & & uzb&C &Uzbek & 25.2 & 64.5 & 39.3 & & bcl&L &Central
   Bikol& 10.2 & 79.8 & 69.6 \\ \cline{1-1} \cline{3-8} \cline{10-15} \rule{0pt}{2.5ex}
    \multirow{5}{*}{ \makecell{\rotatebox{90}{low end}}}    && dtp&L &Kadazan Dusun& 5.6 & 21.1 & 15.5 & & xav&L& Xavánte
    & 2.2 & 5.0 & 2.8 \\
    && kab&L &Kabyle & 3.7 & 16.4 & 12.7 & & mau&L& Huautla
    Mazatec & 2.4 & 3.6 & 1.2 \\
    && pam&L &Pampanga & 4.8 & 11.0 & 6.2 & & ahk&L& Akha& 3.0 & 3.2 & 0.2 \\
    && lvs&L &Standard Latvian & 73.4 & 76.9 & 3.5 & & aln&L&
    Gheg Albanian& 67.8 & 67.6 & -0.2 \\
    && nob&L &Bokmål& 93.5 & 95.7 & 2.2 & & nob&L& Bokmål&
    82.8 & 79.2 & -3.6 \\ \hline \hline \rule{0pt}{2.5ex}
    \multirow{5}{*}{ \makecell{\rotatebox{90}{high end}}}    &\multirow{10}{*}{ \rotatebox{90}{NER} } & div&T &Dhivehi & 0.0 &
  50.9 & 50.9 & \multirow{10}{*}{ \rotatebox{90}{POS }} &
  mlt&L& Maltese & 21.3 & 80.3 & 59.0 \\
    && che&C &Chechen& 15.3 & 61.2 & 45.9 & & sah&C& Yakut& 21.9 & 76.9 & 55.0 \\
    && mri&L &Maori & 16.0 & 58.9 & 42.9 & & sme&L &Northern Sami& 29.6 & 73.6 & 44.1 \\
    && nan&L &Min Nan& 42.3 & 84.9 & 42.6 & & yor&L &Yoruba& 22.8 & 64.2 & 41.4 \\
    && tgk&C &Tajik& 26.3 & 66.4 & 40.0 & & quc&L &K'iche' &
    28.5 & 64.1 & 35.6 \\ \cline{1-1} \cline{3-8} \cline{10-15} \rule{0pt}{2.5ex}
    \multirow{5}{*}{ \makecell{\rotatebox{90}{low end}}}    && zea&L &Zeeuws & 68.1 & 67.3 & -0.8 & & lzh&H &Literary Chinese& 11.7 & 18.4 & 6.7 \\ 
    && vol&L &Volapük& 60.0 & 59.0 & -1.0 & & nap&L &Neapolitan& 47.1 & 50.0 & 2.9 \\
    && min&L &Minangkabau& 42.3 & 40.4 & -1.8 & & hyw&A& Western Armenian& 79.1 & 81.1 & 2.0 \\
    && wuu&H &Wu Chinese & 28.9 & 23.9 & -5.0 & & kmr&L &Northern
    Kurdish & 73.5 & 75.2 & 1.7 \\
    && lzh&H &Literary Chinese& 15.7 & 10.3 & -5.4 & & aln&L&
    Gheg Albanian& 54.7 & 51.2 & -3.5 
    \end{tabular}
    \caption{Results
 for five 
\unseenlanguagescripts each with the largest (high end) and
smallest (low end) gain
\modelname vs.\  \xlmrb for four tasks.
\modelname's gain over \xlmrb is  large at the
high end and small or slightly negative  at the
low end.
L = Latin,
C = Cyrillic,
H = Hani,
A = Armenian,
T = Thaana
}
\tablabel{analysis_lang}
\end{table*}

\subsection{Analysis across \languagescriptscap{}}
\seclabel{analysislanguagescript}
To analyze the effect of
\languagescripts, we select
five \unseenlanguagescripts
each
with the largest and smallest gain when comparing
\modelname vs.\ \xlmrb
for SentRetr and sequence labeling.

Table~\ref{tab:analysis_lang}
shows that \modelname
improves languages with scripts not
covered by XLM-R
(e.g.,
div/Dhivehi,
Thaana script, see \secref{coverage}) by a large margin  since XLM-R  simply regards the uncovered scripts as
unknown tokens and cannot compute meaningful representations
for the input. The large amount of
data we collected in \corpusnameclean
also contributes to the improvement for \unseenlanguages,
e.g., for  tat\_Cyrl (Tatar) in SentRetr Tatoeba
and mlt\_Latn (Maltese) in POS. See \secref{AmountoftrainingDataeffect}
for a  detailed analysis of the effect of corpus size.

On the other hand, \modelname achieves just comparable or
even worse results for some \languagescripts.
We see at least three explanations.
(i) As discussed in \secref{coverage},
some \unseenlanguages (e.g.,
nob/Norwegian Bokmal) are close to
a \seenlanguage (e.g., nor/Norwegian), so \modelname has no
advantage over \xlmrb.
(ii) A language is at the low end of our corpus size
range (i.e., 30,000 sentences'). Example:
xav\_Latn, Xavánte. (iii) Some languages are completely
distinct from all other languages in \corpusnameclean,
thus without support from any similar language. An example is
mau\_Latn, Huautla Mazatec. \modelname
has a much harder time learning good representations in
these cases.

\begin{table}[t]
  \centering
  \small
  \def\tablesep{0.1cm}
    \begin{tabular}{cc|rrr}
    lang-script &  & \tabmc{\xlmrb} & \modelname & gain \\ \midrule
    uig\_Arab & \seenlanguagesadj & 45.8 & 56.2 & 10.4 \\
    uig\_Latn & \unseenlanguagesadj & 9.8 & 62.8 & 53.0 \\ \hline \rule{0pt}{2ex}
    hin\_Deva & \seenlanguagesadj & 67.0 & 76.6 & 9.6 \\
    hin\_Latn & \unseenlanguagesadj& 13.6 & 43.2 & 29.6 \\ \hline \rule{0pt}{2ex}
    uzb\_Latn & \seenlanguagesadj & 54.8 & 67.6 & 12.8 \\
    uzb\_Cyrl & \unseenlanguagesadj& 6.2 & 78.8 & 72.6 \\ \hline \rule{0pt}{2ex}
    kaa\_Cyrl & \unseenlanguagesadj& 17.6 & 73.8 & 56.2 \\
    kaa\_Latn & \unseenlanguagesadj& 9.2 & 43.4 & 34.2 \\ \hline \rule{0pt}{2ex}
    kmr\_Cyrl & \unseenlanguagesadj& 4.0 & 42.4 & 38.4 \\
    kmr\_Latn & \unseenlanguagesadj& 35.8 & 63.0 & 27.2 \\ \hline \rule{0pt}{2ex}
    tuk\_Cyrl & \unseenlanguagesadj& 13.6 & 65.0 & 51.4 \\
    tuk\_Latn & \unseenlanguagesadj& 9.6 & 66.2 & 56.6 
    \end{tabular}
\caption{Sentence Retrieval Bible performance
of  \modelname and \xlmrb for six languages with two
scripts:
Uighur (uig), Hindi (hin), Uzbek (uzb), Kara-Kalpak (kaa),
Northern Kurdish (kmr), Turkmen (tuk).
\modelname clearly outperforms \xlmrb
with large differences for \unseenlanguagescripts.}
\tablabel{analysis_script}
\end{table}

\subsection{Languages with Multiple Scripts}

Table~\ref{tab:analysis_script} compares
SentRetr performance
\xlmrb vs.\ \modelname for six languages with two scripts.
Unsurprisingly, XLM-R performs much better for a
language-script  it was pretrained on (``head'') than on one
that it was not (``tail'').
We can improve the
performance of a language, even
surpassing the language-script covered by XLM-R,
if we collect
enough data for its script not covered by XLM-R.
For languages with two scripts not covered by XLM-R, the
performance is better for the script for which we collect a larger corpus. For
example, kaa\_Cyrl (Kara-Kalpak) has about three times as much data as
kaa\_Latn.
This explains why kaa\_Cyrl outperforms
kaa\_Latn by 30\%. 

\citet{dufter-schutze-2020-identifying} found that, after
training a multilingual model with two scripts for English
(natural English and ``fake English''), the model performed
well at zero-shot transfer if the capacity of the model was
of the right size (i.e., not too small, not too large).
Our experiments with real data show the complexity of
the issue: even if there is a ``right'' size for an LLM that
supports both full acquisition of languages and multilingual
transfer, this size
is difficult to determine and it
may be different for different language
pairs in a large horizontally scaled model like \modelname.

\begin{table}[t]
  \centering
  \small
  \def\tablesep{0.1cm}
    \begin{tabular}{
  @{\hspace{\tablesep}}l@{\hspace{\tablesep}}|
  @{\hspace{\tablesep}}r@{\hspace{\tablesep}}
  @{\hspace{\tablesep}}r@{\hspace{\tablesep}}
  @{\hspace{\tablesep}}r@{\hspace{\tablesep}}
  @{\hspace{\tablesep}}c@{\hspace{\tablesep}}
  @{\hspace{\tablesep}}c@{\hspace{\tablesep}}
}
    family & $|L_G|$&$|L_X|$ & \xlmrb & \modelname & gain \\ \midrule
    indo1319 & 91&50 & 41.5 & 61.4 & 19.9 \\
    atla1278 & 69&2 & 5.5 & 45.2 & 39.6 \\
    aust1307 & 53&6 & 13.7 & 47.0 & 33.2 \\
    turk1311 & 22&7 & 20.1 & 62.9 & 42.8 \\
    sino1245 & 22&2 & 7.6 & 38.9 & 31.3 \\
    maya1287 & 15&0 & 3.8 & 20.3 & 16.4 \\
    afro1255 & 12&5 & 13.0 & 34.3 & 21.4 
    \end{tabular}

\caption{Average Sentence Retrieval Bible performance
of  \modelname and \xlmrb for
seven language families.
The difference in coverage of a family by \modelname vs.\
\xlmrb is partially predictive of the
performance difference.
$|L_G|$/$|L_X|$:
number of language-scripts from family covered by \modelname/XLM-R.
}
\tablabel{analysis_family}
\end{table}

\subsection{Analysis across Language Families}
Table~\ref{tab:analysis_family}
compares  SentRetr  performance
 \modelname vs.\ \xlmrb for
seven language families that have  ten
or more language-scripts in \corpusnameclean.
We assign languages to families based on
Glottolog.\footnote{\url{http://glottolog.org/glottolog/family}}
Generally, XLM-R has better  performance
the more \languagescripts from a language
family are represented
in its training data; e.g., performance is better for 
indo1319 and worse for
maya1287.
The results suggest that
\modelname's improvement over XLM-R is the larger,
the
better our training corpus \corpusnameclean's coverage is of
a family.

\subsection{Effect of Amount of Training Data }
\seclabel{AmountoftrainingDataeffect}
We examine
correlation between
pretraining corpus size
and \modelname zero-shot performance.
We focus on SentRetr Bible
(\secref{tasksentenceretrieval})
since it
supports the  most \seenlanguagesadj and \unseenlanguages.
We find that Pearson's $r = .34$, i.e., corpus size and
performance are moderately, but clearly correlated.
We suspect that the correlation is not larger because,
in addition to corpus size of language $l$ itself, corpus
size of languages closely related to $l$ is also an
important factor
(see \secref{analysislanguagescript} for a similar finding for
Norwegian).
We therefore also compute Pearson's $r$
between (i) performance of language $l$ on SentRetr Bible and (ii)
joint corpus size of $l$ and its $k$ nearest neighbors
(according to perplexity divergence, \secref{ngramlms}).
In this case, Pearson's $r =.44$ (for both $k=3$ and $k=4$), indicating that the corpus
size of nearest neighbor languages does play a role.

\begin{table}
  \small
	\centering
	\def\tablesep{0.1cm}
\begin{tabular}{
  @{\hspace{\tablesep}}l@{\hspace{\tablesep}}|
  @{\hspace{\tablesep}}c@{\hspace{\tablesep}}
  @{\hspace{\tablesep}}c@{\hspace{\tablesep}}
}
  lang-script & \modelnamejustone & \modelname \\
  \midrule
  rug\_Latn, Roviana & \textbf{51.0} & 49.0  \\
  yan\_Latn, Mayangna/Sumo & \textbf{46.4} & 31.8 \\
  wbm\_Latn, Wa/Va & \textbf{49.6} & 46.4 \\
  \midrule
  ctd\_Latn, Tedim Chin& 47.4 & \textbf{59.4} \\
  quh\_Latn, Southern Quechua& 33.4 & \textbf{56.2} \\
  tat\_Cyrl, Tatar & 58.8 & \textbf{67.2} 
\end{tabular}
\caption{Performance on Sentence Retrieval Bible of continued pretraining on just one
  \languagescript (\modelnamejustone) vs.\ on \corpusnameclean
  (\modelname). \modelname underperforms on the top three
  and outperforms on the bottom three.
Our explanation is that the second group is supported by
  closely related languages in \corpusnameclean; e.g.,
for Southern Quechua (quh),
\modelname
  also covers closely related Cuzco Quechua (quz). For the
  first group this is not the case; e.g., the Wa language
  (wbm) has no close relative in \corpusnameclean.}
\tablabel{single_lm}
\end{table}

\subsection{Support through Related Languages}
\seclabel{supportthroughrelatedlanguages}
Building on \secref{AmountoftrainingDataeffect}, there is
another way we can investigate the positive effect of
closely related languages on performance:
We can compare
performance (again on SentRetr Bible) of continued pretraining on just one
  language (we refer to this model as \modelnamejustone) vs.\ on
all \numberlanguagesclean languages 
represented in \corpusnameclean
  (i.e., \modelname).
Table~\ref{tab:single_lm}
presents results for six \languagescripts selected from various
language families and
suggests that some languages do
  not receive support from related languages (top three). In that
  case, \modelnamejustone can fully concentrate on learning
  the isolated language and does better
  than \corpusnameclean.
  Other languages (bottom three) do receive support from related languages.
  For example,
Southern Quechua (quh) seems to receive support in
\modelname from closely related Cuzco Quechua (quz),
resulting in \modelname outperforming \modelnamejustone.

\section{Conclusion and Future Work}
We collect and data-clean \corpusnameclean, a large corpus
of hundreds of usually neglected
tail (i.e., long-tail) languages
and create \modelname, an LLM
that is trained on \corpusnameclean and covers
these languages. We evaluate \modelname
on \numberoftasks tasks
that allow us to evaluate almost all languages.
We observe large improvements for
both \seenlanguagesadj and \unseenlanguages
compared to XLM-R.
Our analysis shows that no single
factor fully explains the quality of the representation of a language in
a multilingual model. Rather, a combination of factors
is important, including
corpus size, script, ``help'' from related languages and the
total capacity of the model.

This work is the first to create a language model on a
dataset of several hundreds of gigabytes and to make it
publicly available for such a large and diverse number of
low-resource languages. In future research, we would like to
train larger models to further investigate the effect of
model size, distill highly multilingual models for
resource-efficient deployment, explore alternatives to
continued pretraining and use models for more
\unseenlanguagesadj language downstream tasks.

\section*{Limitations}
(1) We did not perform any comprehensive hyperparameter search, which would have further consolidated our results.
This decision was made due to the high cost of training multiple models.
(2) Compared to current very large models, \modelname is comparatively small.
(3) Although we have tried to minimize the amount of noise in our data, some noise is still present.

\section*{Ethics Statement}
There are two issues worth mentioning in regards to this project.
First, it was not feasible for us to thoroughly examine the 
content of the data for all languages, thus we cannot confirm the 
absence of discrimination based on factors such as race or sexuality. 
The data was solely utilized as a textual corpus, and the content should 
not be interpreted as an endorsement by our team. 
If the model is subsequently utilized for generation, 
it is possible that the training data may be reflected in the generated output. 
However, addressing potential biases within the data is an area for future research. 
Second, it is important to note that while the \dsources utilized 
in this study do not explicitly prohibit the reuse of data for 
research purposes, some sources do have copyright statements indicating 
that such use is permissible while others do not. 
Additionally, certain sources prohibit the redistribution of data. 
As such, data from these sources is omitted from the published 
version of \corpusnameraw.

\section*{Acknowledgements}
We would like to thank Renhao Pei, Yihong Liu, Verena Blaschke, and the anonymous reviewers. This work was funded by the European Research Council (grants \#740516 and \#758969) and EU's Horizon Europe Research and Innovation Actions (UTTER, contract 101070631).

\bibliography{anthology,custom}
\bibliographystyle{acl_natbib}

\appendix

\section{N-grams LMs and Language Divergence}\label{sec:appendix_ngrams}
\noindent \textbf{Perplexity and Language Divergence.}
Perplexity measures how well a model predicts a sample test data. Assuming a test data contains sequences of characters $S$ = $ch_1, ch_2, \cdots, ch_T$, perplexity ($\mathcal{PP}$) of $S$ given an \ngram  character level language model $M$ is computed as follows: 
\begin{equation}
\mathcal{PP}(S, M)=\sqrt[T]{\prod_{t=1}^T \frac{1}{\mathbb{P}\left(c h_t \mid c h_1^{t-1}\right)}}
\end{equation}
where $\mathbb{P}\left(ch_t \mid ch_1^{t-1}\right)$ is computed as by dividing the observed frequency ($C$) of $ch_1^{t-1} ch_i$ by the observed frequency of $ch_1^{t-1}$ in $M$ training data:
\begin{equation}
\mathbb{P}\left(ch_t \mid ch_1^{t-1}\right)=\frac{C\left(c h_1^{t-1} c h_t\right)}{C\left(ch_1^{t-1}\right)}
\end{equation}
Given the definition of perplexity, we can determine how well a trained language model on language $L_1$ predicts the test text of language $L_2$ and vice-versa.
The divergence between two languages is computed with the maximum of the perplexity values in both directions. Two reasons lead to the use of $\max$: first, a symmetrical divergence is required, and second, languages differ in their complexity, so one direction of computing perplexity may result in a much lower perplexity than another. Thus, comparing perplexity results becomes difficult. As an example, the Kuanua language (ksd\_Latn) has short words and a simple structure, which results in $3-$gram models getting lower perplexity on its text compared to other languages. The lower the perplexity the smaller the divergence between languages. The divergence ($\mathcal{D}$) between language $L_i$ and $L_j$ with trained language models of $M_{L_z}$ and test texts of $S_{L_z}$, where $L_z$ is the corresponding language, computed as follows:
\begin{equation}\eqlabel{pp-dist}
\resizebox{0.89\hsize}{!}{$\mathcal{D}_{L_i, L_j}= \max \big(\mathcal{PP}(S_{L_i}, M_{L_j}), \mathcal{PP}(S_{L_j}, M_{L_i}) \big)$}
\end{equation}

\noindent \textbf{Runs and Data.} 
The data used to train and test the character level \ngram models is the same data used for the training and testing of the \modelname{}. The training of the models was limited to $100,000$ sentences' per language-script. We use KenLM library~\cite{heafield-2011-kenlm} to build \ngram models. This library uses an interpolated modified Kneser-Ney smoothing for estimating the unseen \ngram{s}. Our evaluation has been performed over 7 \ngram models ($3 \leq n \leq 9$).

\noindent \textbf{Baseline and Evaluation.}
Language family trees were used as a baseline for evaluating the divergence measures of the proposed approach. We obtained language family tree data from Ethnologue online version \cite{ethnologue}. For each language, the family tree follows the general order from largest typological language family group to smallest.
There is only one family tree for each language in the baseline data. Nodes in the family tree represent typological language family groups. Each node only has one parent, so if a node is common in the family tree of two languages, its parent is also common.
We evaluate our perplexity method on the following binary classification task: Do the majority of a language $L_z$’s $k$ nearest neighbors belong to the same typological language family group as $L_z$?
Assuming  languages $L_i$ and $L_j$, with the following family trees:
\begin{align*}
\centering
&T_{L_i}:~\circledBody{1}~\rightarrow~\circledBody{2}~\rightarrow~\circledBody{3}~\rightarrow~\circledBody{4}~\rightarrow~\circledBody{5}~\rightarrow~\circledBody{6} \\
&T_{L_j}:~\circledBody{1}~\rightarrow~\circledBody{2}~\rightarrow~\circledBody{7}~\rightarrow~\circledBody{8}
\end{align*}
These 2 languages belong to the same typological family group with family tree levels of $l \in \{1, 2\}$, but not with family tree levels of $l=3$ and higher.

\noindent \textbf{Result.}
When it comes to language families, the majority of studies only refer to the largest typological language family group (level $l=1$). Here, we also assess our methodology for other levels.
The results of classification accuracy for $3-$gram model, $ k \in \{1, 3, 7, 13, 21 \}$ and $l \in \{1, 2, 3, \max \}$ are shown in Table \ref{tab:langcomp}. 
In cases where the maximum level of a tree is less than the $l$ parameter, the maximum level for that language is used. Languages without a family or no other family member in our data are excluded. We only report the $3-$gram model results as it gets the best results in most configurations among other \ngram models. With increasing $l$, the accuracy decreases, since more languages fall outside the same typological family. As $k$ increases, the accuracy decreases, because languages with faraway neighbors are being included but the number of languages in the language typological group family will remain the same. 
There are times when languages have a lot of loan words from other languages because of geological proximity or historical reasons (e.g, colonization), which makes them similar to the languages they borrowed words from in our method. However they are different when it comes to their typological families and our method fails in these cases.
Aymara (Macrolanguage: aym\_Latn) and Quechua (Macrolanguage: que\_Latn), for example, had a great deal of contact and influence on each other, but they do not belong to the same typological group.
As well, some of the typological families are not that large, which makes our results worse when $k$ increases. This is the case, for instance, of the Tarascan typological family which only has two members.

\begin{table}[ht]
\centering
\resizebox{0.7\linewidth}{!}{
\begin{tabular}{c c c c c} 
\toprule
model & $l$ & $k$ & accuracy (\%)\\
\midrule
$3$-gram & $1$ & $1$ & $84.45$ \\
$3$-gram & $1$ & $3$ & $75.77$ \\
$3$-gram & $1$ & $7$ & $69.08$ \\
$3$-gram & $1$ & $13$ & $62.75$ \\
$3$-gram & $1$ & $21$ & $55.33$ \\
$3$-gram & $2$ & $1$ & $79.75$ \\
$3$-gram & $2$ & $3$ & $67.63$ \\
$3$-gram & $2$ & $7$ & $59.49$ \\
$3$-gram & $2$ & $13$ & $51.36$ \\
$3$-gram & $2$ & $21$ & $42.68$ \\
$3$-gram & $3$ & $1$ & $75.05$ \\
$3$-gram & $3$ & $3$ & $60.22$ \\
$3$-gram & $3$ & $7$ & $49.55$ \\
$3$-gram & $3$ & $13$ & $38.34$ \\
$3$-gram & $3$ & $21$ & $29.84$ \\
$3$-gram & $\max$ & $1$ & $59.31$ \\
$3$-gram & $\max$ & $3$ & $36.89$ \\
$3$-gram & $\max$ & $7$ & $18.81$ \\
$3$-gram & $\max$ & $13$ & $6.87$ \\
$3$-gram & $\max$ & $21$ & $2.89$ \\
\bottomrule
\end{tabular}
}
\caption{Detecting the typological relatedness of language with n-gram divergence: (\qref{pp-dist}); $l$: level of typological language family group; $k$: number of nearest language neighbors.}
\label{tab:langcomp}
\end{table}

\section{Languages}
\label{sec:appendix_languages}
The list of languages used to train \modelname{} 
with the amount of available data for each language is available 
in Tables~\ref{tab:languages1}, ~\ref{tab:languages2} and \ref{tab:languages3}.

\paragraph{On Macrolanguages} The presence of language codes that are supersets of other language codes within datasets is not uncommon~\cite{kreutzer-etal-2022-quality}. This issue becomes more prevalent in extensive collections. Within the ISO 639-3 standard, these languages are referred to as macrolanguages. When confronted with macrolanguages, if it is not feasible to ascertain the specific individual language contained within a dataset, the macrolanguage code is retained. Consequently, it is possible that in \corpusnameraw and \corpusnameclean both the corpora for the macrolanguage and its individual languages have been included.

\begin{table*}
\centering
\resizebox{\textwidth}{!}{
    \begin{tabular}{cccc|cccc|cccc}
    \toprule
        \languagescriptcap & |Sent| & Family & Head & \languagescriptcap & |Sent| & Family & Head & \languagescriptcap & |Sent| & Family & Head \\
    \midrule
        hbs\_Latn & 63411156 & indo1319 & & vec\_Latn & 514240 & indo1319 & & swh\_Latn & 95776 & atla1278 & yes \\
mal\_Mlym & 48098273 & drav1251 & yes & jpn\_Jpan & 510722 & japo1237 & yes & alt\_Cyrl & 95148 & turk1311 & \\
aze\_Latn & 46300705 & & yes & lus\_Latn & 509250 & sino1245 & & rmn\_Grek & 94533 & indo1319 & \\
guj\_Gujr & 45738685 & indo1319 & yes & crs\_Latn & 508755 & indo1319 & & miq\_Latn & 94343 & misu1242 & \\
ben\_Beng & 43514870 & indo1319 & yes & kqn\_Latn & 507913 & atla1278 & & kaa\_Cyrl & 88815 & turk1311 & \\
kan\_Knda & 41836495 & drav1251 & yes & ndo\_Latn & 496613 & atla1278 & & kos\_Latn & 88603 & aust1307 & \\
tel\_Telu & 41580525 & drav1251 & yes & snd\_Arab & 488730 & indo1319 & yes & grn\_Latn & 87568 & & \\
mlt\_Latn & 40654838 & afro1255 & & yue\_Hani & 484700 & sino1245 & & lhu\_Latn & 87255 & sino1245 & \\
fra\_Latn & 39197581 & indo1319 & yes & tiv\_Latn & 483064 & atla1278 & & lzh\_Hani & 86035 & sino1245 & \\
spa\_Latn & 37286756 & indo1319 & yes & kua\_Latn & 473535 & atla1278 & & ajp\_Arab & 83297 & afro1255 & \\
eng\_Latn & 36122761 & indo1319 & yes & kwy\_Latn & 473274 & atla1278 & & cmn\_Hani & 80745 & sino1245 & yes \\
fil\_Latn & 33493255 & aust1307 & yes & hin\_Latn & 466175 & indo1319 & & gcf\_Latn & 80737 & indo1319 & \\
nob\_Latn & 32869205 & indo1319 & & iku\_Cans & 465011 & & & rmn\_Cyrl & 79925 & indo1319 & \\
rus\_Cyrl & 31787973 & indo1319 & yes & kal\_Latn & 462430 & eski1264 & & kjh\_Cyrl & 79262 & turk1311 & \\
deu\_Latn & 31015993 & indo1319 & yes & tdt\_Latn & 459818 & aust1307 & & rng\_Latn & 78177 & atla1278 & \\
tur\_Latn & 29184662 & turk1311 & yes & gsw\_Latn & 449240 & indo1319 & & mgh\_Latn & 78117 & atla1278 & \\
pan\_Guru & 29052537 & indo1319 & yes & mfe\_Latn & 447435 & indo1319 & & xmv\_Latn & 77896 & aust1307 & \\
mar\_Deva & 28748897 & indo1319 & yes & swc\_Latn & 446378 & atla1278 & & ige\_Latn & 77114 & atla1278 & \\
por\_Latn & 27824391 & indo1319 & yes & mon\_Latn & 437950 & mong1349 & & rmy\_Latn & 76991 & indo1319 & \\
nld\_Latn & 25061426 & indo1319 & yes & mos\_Latn & 437666 & atla1278 & & srm\_Latn & 76884 & indo1319 & \\
ara\_Arab & 24524122 & & yes & kik\_Latn & 437228 & atla1278 & & bak\_Latn & 76809 & turk1311 & \\
zho\_Hani & 24143786 & & yes & cnh\_Latn & 436667 & sino1245 & & gur\_Latn & 76151 & atla1278 & \\
ita\_Latn & 23539857 & indo1319 & yes & gil\_Latn & 434529 & aust1307 & & idu\_Latn & 75106 & atla1278 & \\
ind\_Latn & 23018106 & aust1307 & yes & pon\_Latn & 434522 & aust1307 & & yom\_Latn & 74818 & atla1278 & \\
ell\_Grek & 22033282 & indo1319 & yes & umb\_Latn & 431589 & atla1278 & & tdx\_Latn & 74430 & aust1307 & \\
bul\_Cyrl & 21823004 & indo1319 & yes & lvs\_Latn & 422952 & indo1319 & & mzn\_Arab & 73719 & indo1319 & \\
swe\_Latn & 20725883 & indo1319 & yes & sco\_Latn & 411591 & indo1319 & & cfm\_Latn & 70227 & sino1245 & \\
ces\_Latn & 20376340 & indo1319 & yes & ori\_Orya & 410827 & & yes & zpa\_Latn & 69237 & otom1299 & \\
isl\_Latn & 19547941 & indo1319 & yes & arg\_Latn & 410683 & indo1319 & & kbd\_Cyrl & 67914 & abkh1242 & \\
pol\_Latn & 19339945 & indo1319 & yes & kur\_Latn & 407169 & indo1319 & yes & lao\_Laoo & 66966 & taik1256 & yes \\
ron\_Latn & 19190217 & indo1319 & yes & dhv\_Latn & 405711 & aust1307 & & nap\_Latn & 65826 & indo1319 & \\
dan\_Latn & 19174573 & indo1319 & yes & luo\_Latn & 398974 & nilo1247 & & qub\_Latn & 64973 & quec1387 & \\
hun\_Latn & 18800025 & ural1272 & yes & lun\_Latn & 395764 & atla1278 & & oke\_Latn & 64508 & atla1278 & \\
tgk\_Cyrl & 18659517 & indo1319 & & nzi\_Latn & 394247 & atla1278 & & ote\_Latn & 64224 & otom1299 & \\
srp\_Latn & 18371769 & indo1319 & yes & gug\_Latn & 392227 & tupi1275 & & bsb\_Latn & 63634 & aust1307 & \\
fas\_Arab & 18277593 & & yes & bar\_Latn & 387070 & indo1319 & & ogo\_Latn & 61901 & atla1278 & \\
ceb\_Latn & 18149215 & aust1307 & & bci\_Latn & 384059 & atla1278 & & abn\_Latn & 61830 & atla1278 & \\
heb\_Hebr & 18128962 & afro1255 & yes & chk\_Latn & 380596 & aust1307 & & ldi\_Latn & 61827 & atla1278 & \\
hrv\_Latn & 17882932 & indo1319 & yes & roh\_Latn & 377067 & indo1319 & & ayr\_Latn & 61570 & ayma1253 & \\
glg\_Latn & 17852274 & indo1319 & yes & aym\_Latn & 373329 & ayma1253 & & gom\_Deva & 61140 & indo1319 & \\
fin\_Latn & 16730388 & ural1272 & yes & yap\_Latn & 358929 & aust1307 & & bba\_Latn & 61123 & atla1278 & \\
slv\_Latn & 15719210 & indo1319 & yes & ssw\_Latn & 356561 & atla1278 & & aln\_Latn & 60989 & indo1319 & \\
vie\_Latn & 15697827 & aust1305 & yes & quz\_Latn & 354781 & quec1387 & & leh\_Latn & 59944 & atla1278 & \\
mkd\_Cyrl & 14717004 & indo1319 & yes & sah\_Cyrl & 352697 & turk1311 & & ban\_Latn & 59805 & aust1307 & \\
slk\_Latn & 14633631 & indo1319 & yes & tsn\_Latn & 350954 & atla1278 & & ace\_Latn & 59333 & aust1307 & \\
nor\_Latn & 14576191 & indo1319 & yes & lmo\_Latn & 348135 & indo1319 & & pes\_Arab & 57511 & indo1319 & yes \\
est\_Latn & 13600579 & & yes & ido\_Latn & 331239 & arti1236 & & skg\_Latn & 57228 & aust1307 & \\
ltz\_Latn & 12997242 & indo1319 & & abk\_Cyrl & 321578 & abkh1242 & & ary\_Arab & 56933 & afro1255 & \\
eus\_Latn & 12775959 & & yes & zne\_Latn & 318871 & atla1278 & & hus\_Latn & 56176 & maya1287 & \\
lit\_Latn & 12479626 & indo1319 & yes & quy\_Latn & 311040 & quec1387 & & glv\_Latn & 55641 & indo1319 & \\
kaz\_Cyrl & 12378727 & turk1311 & yes & kam\_Latn & 310659 & atla1278 & & fat\_Latn & 55609 & atla1278 & \\
lav\_Latn & 12143980 & indo1319 & yes & bbc\_Latn & 310420 & aust1307 & & frr\_Latn & 55254 & indo1319 & \\
bos\_Latn & 11014744 & indo1319 & yes & vol\_Latn & 310399 & arti1236 & & mwn\_Latn & 54805 & atla1278 & \\
epo\_Latn & 8737198 & arti1236 & yes & wal\_Latn & 309873 & gong1255 & & mai\_Deva & 54687 & indo1319 & \\
cat\_Latn & 8648271 & indo1319 & yes & uig\_Arab & 307302 & turk1311 & yes & dua\_Latn & 53392 & atla1278 & \\
tha\_Thai & 7735209 & taik1256 & yes & vmw\_Latn & 306899 & atla1278 & & dzo\_Tibt & 52732 & sino1245 & \\
ukr\_Cyrl & 7462046 & indo1319 & yes & kwn\_Latn & 305362 & atla1278 & & ctd\_Latn & 52135 & sino1245 & \\
tgl\_Latn & 7411064 & aust1307 & yes & pam\_Latn & 303737 & aust1307 & & nnb\_Latn & 52041 & atla1278 & \\
sin\_Sinh & 7293178 & indo1319 & yes & seh\_Latn & 300243 & atla1278 & & sxn\_Latn & 51749 & aust1307 & \\
gle\_Latn & 7225513 & indo1319 & yes & tsc\_Latn & 298442 & atla1278 & & mps\_Latn & 50645 & tebe1251 & \\
hin\_Deva & 7046700 & indo1319 & yes & nyk\_Latn & 297976 & atla1278 & & mny\_Latn & 50581 & atla1278 & \\
kor\_Hang & 6468444 & kore1284 & yes & kmb\_Latn & 296269 & atla1278 & & gkp\_Latn & 50549 & mand1469 & \\
ory\_Orya & 6266475 & indo1319 & & zai\_Latn & 277632 & otom1299 & & kat\_Latn & 50424 & kart1248 & \\
urd\_Arab & 6009594 & indo1319 & yes & gym\_Latn & 274512 & chib1249 & & bjn\_Latn & 49068 & aust1307 & \\
swa\_Latn & 5989369 & & yes & bod\_Tibt & 273489 & sino1245 & & acr\_Latn & 48886 & maya1287 & \\
sqi\_Latn & 5526836 & indo1319 & yes & nde\_Latn & 269931 & atla1278 & & dtp\_Latn & 48468 & aust1307 & \\
bel\_Cyrl & 5319675 & indo1319 & yes & fon\_Latn & 268566 & atla1278 & & lam\_Latn & 46853 & atla1278 & \\
afr\_Latn & 5157787 & indo1319 & yes & ber\_Latn & 264426 & & & bik\_Latn & 46561 & & \\
nno\_Latn & 4899103 & indo1319 & & nbl\_Latn & 259158 & atla1278 & & poh\_Latn & 46454 & maya1287 & \\
tat\_Cyrl & 4708088 & turk1311 & & kmr\_Latn & 256677 & indo1319 & & phm\_Latn & 45862 & atla1278 & \\
    \bottomrule
    \end{tabular}
}
    \caption{List of languages used to train \modelname{} (Part I).}
    \tablabel{languages1}
\end{table*}

\begin{table*}
\centering
\resizebox{\textwidth}{!}{
    \begin{tabular}{cccc|cccc|cccc}
    \toprule
        \languagescriptcap & |Sent| & Family & Head & \languagescriptcap & |Sent| & Family & Head & \languagescriptcap & |Sent| & Family & Head \\
    \midrule
        ast\_Latn & 4683554 & indo1319 & & guc\_Latn & 249044 & araw1281 & & hrx\_Latn & 45716 & indo1319 & \\
mon\_Cyrl & 4616960 & mong1349 & yes & mam\_Latn & 248348 & maya1287 & & quh\_Latn & 45566 & quec1387 & \\
hbs\_Cyrl & 4598073 & indo1319 & & nia\_Latn & 247406 & aust1307 & & hyw\_Cyrl & 45379 & indo1319 & \\
hau\_Latn & 4368483 & afro1255 & yes & nyn\_Latn & 241992 & atla1278 & & rue\_Cyrl & 45369 & indo1319 & \\
sna\_Latn & 4019596 & atla1278 & & cab\_Latn & 240101 & araw1281 & & eml\_Latn & 44630 & indo1319 & \\
msa\_Latn & 3929084 & & yes & top\_Latn & 239232 & toto1251 & & acm\_Arab & 44505 & afro1255 & \\
som\_Latn & 3916769 & afro1255 & yes & tog\_Latn & 231969 & atla1278 & & tob\_Latn & 44473 & guai1249 & \\
srp\_Cyrl & 3864091 & indo1319 & yes & mco\_Latn & 231209 & mixe1284 & & ach\_Latn & 43974 & nilo1247 & \\
mlg\_Latn & 3715802 & & yes & tzh\_Latn & 230706 & maya1287 & & vep\_Latn & 43076 & ural1272 & \\
zul\_Latn & 3580113 & atla1278 & & pms\_Latn & 227748 & indo1319 & & npi\_Deva & 43072 & indo1319 & \\
arz\_Arab & 3488224 & afro1255 & & wuu\_Hani & 224088 & sino1245 & & tok\_Latn & 42820 & arti1236 & \\
nya\_Latn & 3409030 & atla1278 & & plt\_Latn & 220413 & aust1307 & & sgs\_Latn & 42467 & indo1319 & \\
tam\_Taml & 3388255 & drav1251 & yes & yid\_Hebr & 220214 & indo1319 & yes & lij\_Latn & 42447 & indo1319 & \\
hat\_Latn & 3226932 & indo1319 & & ada\_Latn & 219427 & atla1278 & & myv\_Cyrl & 42147 & ural1272 & \\
uzb\_Latn & 3223485 & turk1311 & yes & iba\_Latn & 213615 & aust1307 & & tih\_Latn & 41873 & aust1307 & \\
sot\_Latn & 3205510 & atla1278 & & kek\_Latn & 209932 & maya1287 & & tat\_Latn & 41640 & turk1311 & \\
uzb\_Cyrl & 3029947 & turk1311 & & koo\_Latn & 209375 & atla1278 & & lfn\_Latn & 41632 & arti1236 & \\
cos\_Latn & 3015055 & indo1319 & & sop\_Latn & 206501 & atla1278 & & cgg\_Latn & 41196 & atla1278 & \\
als\_Latn & 2954874 & indo1319 & & kac\_Latn & 205542 & sino1245 & & ful\_Latn & 41188 & atla1278 & \\
amh\_Ethi & 2862985 & afro1255 & yes & qvi\_Latn & 205447 & quec1387 & & gor\_Latn & 41174 & aust1307 & \\
sun\_Latn & 2586011 & aust1307 & yes & cak\_Latn & 204472 & maya1287 & & ile\_Latn & 40984 & arti1236 & \\
war\_Latn & 2584810 & aust1307 & & kbp\_Latn & 202877 & atla1278 & & ium\_Latn & 40683 & hmon1336 & \\
div\_Thaa & 2418687 & indo1319 & & ctu\_Latn & 201662 & maya1287 & & teo\_Latn & 40203 & nilo1247 & \\
yor\_Latn & 2392359 & atla1278 & & kri\_Latn & 201087 & indo1319 & & kia\_Latn & 40035 & atla1278 & \\
fao\_Latn & 2365271 & indo1319 & & mau\_Latn & 199134 & otom1299 & & crh\_Cyrl & 39985 & turk1311 & \\
uzn\_Cyrl & 2293672 & turk1311 & & scn\_Latn & 199068 & indo1319 & & crh\_Latn & 39896 & turk1311 & \\
smo\_Latn & 2290439 & aust1307 & & tyv\_Cyrl & 198649 & turk1311 & & enm\_Latn & 39809 & indo1319 & \\
bak\_Cyrl & 2264196 & turk1311 & & ina\_Latn & 197315 & arti1236 & & sat\_Olck & 39614 & aust1305 & \\
ilo\_Latn & 2106531 & aust1307 & & btx\_Latn & 193701 & aust1307 & & mad\_Latn & 38993 & aust1307 & \\
tso\_Latn & 2100708 & atla1278 & & nch\_Latn & 193129 & utoa1244 & & cac\_Latn & 38812 & maya1287 & \\
mri\_Latn & 2046850 & aust1307 & & ncj\_Latn & 192962 & utoa1244 & & hnj\_Latn & 38611 & hmon1336 & \\
hmn\_Latn & 1903898 & & & pau\_Latn & 190529 & aust1307 & & ksh\_Latn & 38130 & indo1319 & \\
asm\_Beng & 1882353 & indo1319 & yes & toj\_Latn & 189651 & maya1287 & & ikk\_Latn & 38071 & atla1278 & \\
hil\_Latn & 1798875 & aust1307 & & pcm\_Latn & 187594 & indo1319 & & sba\_Latn & 38040 & cent2225 & \\
nso\_Latn & 1619354 & atla1278 & & dyu\_Latn & 186367 & mand1469 & & zom\_Latn & 37013 & sino1245 & \\
ibo\_Latn & 1543820 & atla1278 & & kss\_Latn & 185868 & atla1278 & & bqc\_Latn & 36881 & mand1469 & \\
kin\_Latn & 1521612 & atla1278 & & afb\_Arab & 183694 & afro1255 & & bim\_Latn & 36835 & atla1278 & \\
hye\_Armn & 1463123 & indo1319 & yes & urh\_Latn & 182214 & atla1278 & & mdy\_Ethi & 36370 & gong1255 & \\
oci\_Latn & 1449128 & indo1319 & & quc\_Latn & 181559 & maya1287 & & bts\_Latn & 36216 & aust1307 & \\
lin\_Latn & 1408460 & atla1278 & & new\_Deva & 181427 & sino1245 & & gya\_Latn & 35902 & atla1278 & \\
tpi\_Latn & 1401844 & indo1319 & & yao\_Latn & 179965 & atla1278 & & ajg\_Latn & 35631 & atla1278 & \\
twi\_Latn & 1400979 & atla1278 & & ngl\_Latn & 178498 & atla1278 & & agw\_Latn & 35585 & aust1307 & \\
kir\_Cyrl & 1397566 & turk1311 & yes & nyu\_Latn & 177483 & atla1278 & & kom\_Cyrl & 35249 & ural1272 & \\
pap\_Latn & 1360138 & indo1319 & & kab\_Latn & 176015 & afro1255 & & knv\_Latn & 35196 & & \\
nep\_Deva & 1317291 & indo1319 & yes & tuk\_Cyrl & 175769 & turk1311 & & giz\_Latn & 35040 & afro1255 & \\
azj\_Latn & 1315834 & turk1311 & & xmf\_Geor & 174994 & kart1248 & & hui\_Latn & 34926 & nucl1709 & \\
bcl\_Latn & 1284493 & aust1307 & & ndc\_Latn & 174305 & atla1278 & & kpg\_Latn & 34900 & aust1307 & \\
xho\_Latn & 1262364 & atla1278 & yes & san\_Deva & 165616 & indo1319 & yes & zea\_Latn & 34426 & indo1319 & \\
cym\_Latn & 1244783 & indo1319 & yes & nba\_Latn & 163485 & atla1278 & & aoj\_Latn & 34349 & nucl1708 & \\
gaa\_Latn & 1222307 & atla1278 & & bpy\_Beng & 162838 & indo1319 & & csy\_Latn & 34126 & sino1245 & \\
ton\_Latn & 1216118 & aust1307 & & ncx\_Latn & 162558 & utoa1244 & & azb\_Arab & 33758 & turk1311 & yes \\
tah\_Latn & 1190747 & aust1307 & & qug\_Latn & 162500 & quec1387 & & csb\_Latn & 33743 & indo1319 & \\
lat\_Latn & 1179913 & indo1319 & yes & rmn\_Latn & 162069 & indo1319 & & tpm\_Latn & 33517 & atla1278 & \\
srn\_Latn & 1172349 & indo1319 & & cjk\_Latn & 160645 & atla1278 & & quw\_Latn & 33449 & quec1387 & \\
ewe\_Latn & 1161605 & atla1278 & & arb\_Arab & 159884 & afro1255 & yes & rmy\_Cyrl & 33351 & indo1319 & \\
bem\_Latn & 1111969 & atla1278 & & kea\_Latn & 158047 & indo1319 & & ixl\_Latn & 33289 & maya1287 & \\
efi\_Latn & 1082621 & atla1278 & & mck\_Latn & 157521 & atla1278 & & mbb\_Latn & 33240 & aust1307 & \\
bis\_Latn & 1070170 & indo1319 & & arn\_Latn & 155882 & arau1255 & & pfl\_Latn & 33148 & indo1319 & \\
orm\_Latn & 1067699 & & yes & pdt\_Latn & 155485 & indo1319 & & pcd\_Latn & 32867 & indo1319 & \\
haw\_Latn & 1062491 & aust1307 & & her\_Latn & 154827 & atla1278 & & tlh\_Latn & 32863 & arti1236 & \\
hmo\_Latn & 1033636 & pidg1258 & & gla\_Latn & 152563 & indo1319 & yes & suz\_Deva & 32811 & sino1245 & \\
kat\_Geor & 1004297 & kart1248 & yes & kmr\_Cyrl & 151728 & indo1319 & & gcr\_Latn & 32676 & indo1319 & \\
pag\_Latn & 983637 & aust1307 & & mwl\_Latn & 150054 & indo1319 & & jbo\_Latn & 32619 & arti1236 & \\
loz\_Latn & 964418 & atla1278 & & nav\_Latn & 147702 & atha1245 & & tbz\_Latn & 32264 & atla1278 & \\
fry\_Latn & 957422 & indo1319 & yes & ksw\_Mymr & 147674 & sino1245 & & bam\_Latn & 32150 & mand1469 & \\
mya\_Mymr & 945180 & sino1245 & yes & mxv\_Latn & 147591 & otom1299 & & prk\_Latn & 32085 & aust1305 & \\
nds\_Latn & 944715 & indo1319 & & hif\_Latn & 147261 & indo1319 & & jam\_Latn & 32048 & indo1319 & \\
run\_Latn & 943828 & atla1278 & & wol\_Latn & 146992 & atla1278 & & twx\_Latn & 32028 & atla1278 & \\
    \bottomrule
    \end{tabular}
}
    \caption{List of languages used to train \modelname{} (Part II).}
    \tablabel{languages2}
\end{table*}

\begin{table*}
\centering
\resizebox{\textwidth}{!}{
    \begin{tabular}{cccc|cccc|cccc}
    \toprule
        \languagescriptcap & |Sent| & Family & Head & \languagescriptcap & |Sent| & Family & Head & \languagescriptcap & |Sent| & Family & Head \\
    \midrule
        pnb\_Arab & 899895 & indo1319 & & sme\_Latn & 146803 & ural1272 & & nmf\_Latn & 31997 & sino1245 & \\
rar\_Latn & 894515 & aust1307 & & gom\_Latn & 143937 & indo1319 & & caq\_Latn & 31903 & aust1305 & \\
fij\_Latn & 887134 & aust1307 & & bum\_Latn & 141673 & atla1278 & & rop\_Latn & 31889 & indo1319 & \\
wls\_Latn & 882167 & aust1307 & & mgr\_Latn & 138953 & atla1278 & & tca\_Latn & 31852 & ticu1244 & \\
ckb\_Arab & 874441 & indo1319 & & ahk\_Latn & 135068 & sino1245 & & yan\_Latn & 31775 & misu1242 & \\
ven\_Latn & 860249 & atla1278 & & kur\_Arab & 134160 & indo1319 & & xav\_Latn & 31765 & nucl1710 & \\
zsm\_Latn & 859947 & aust1307 & yes & bas\_Latn & 133436 & atla1278 & & bih\_Deva & 31658 & & \\
chv\_Cyrl & 859863 & turk1311 & & bin\_Latn & 133256 & atla1278 & & cuk\_Latn & 31612 & chib1249 & \\
lua\_Latn & 854359 & atla1278 & & tsz\_Latn & 133251 & tara1323 & & kjb\_Latn & 31471 & maya1287 & \\
que\_Latn & 838486 & & & sid\_Latn & 130406 & afro1255 & & hne\_Deva & 31465 & indo1319 & \\
sag\_Latn & 771048 & atla1278 & & diq\_Latn & 128908 & indo1319 & & wbm\_Latn & 31394 & aust1305 & \\
guw\_Latn & 767918 & atla1278 & & srd\_Latn & 127064 & & & zlm\_Latn & 31345 & aust1307 & \\
bre\_Latn & 748954 & indo1319 & yes & tcf\_Latn & 126050 & otom1299 & & tui\_Latn & 31161 & atla1278 & \\
toi\_Latn & 745385 & atla1278 & & bzj\_Latn & 124958 & indo1319 & & ifb\_Latn & 30980 & aust1307 & \\
pus\_Arab & 731992 & indo1319 & yes & udm\_Cyrl & 121705 & ural1272 & & izz\_Latn & 30894 & atla1278 & \\
che\_Cyrl & 728201 & nakh1245 & & cce\_Latn & 120636 & atla1278 & & rug\_Latn & 30857 & aust1307 & \\
pis\_Latn & 714783 & indo1319 & & meu\_Latn & 120273 & aust1307 & & aka\_Latn & 30704 & atla1278 & \\
kon\_Latn & 685194 & & & chw\_Latn & 119751 & atla1278 & & pxm\_Latn & 30698 & book1242 & \\
oss\_Cyrl & 683517 & indo1319 & & cbk\_Latn & 118789 & indo1319 & & kmm\_Latn & 30671 & sino1245 & \\
hyw\_Armn & 679819 & indo1319 & & ibg\_Latn & 118733 & aust1307 & & mcn\_Latn & 30666 & afro1255 & \\
iso\_Latn & 658789 & atla1278 & & bhw\_Latn & 117381 & aust1307 & & ifa\_Latn & 30621 & aust1307 & \\
nan\_Latn & 656389 & sino1245 & & ngu\_Latn & 116851 & utoa1244 & & dln\_Latn & 30620 & sino1245 & \\
lub\_Latn & 654390 & atla1278 & & nyy\_Latn & 115914 & atla1278 & & ext\_Latn & 30605 & indo1319 & \\
lim\_Latn & 652078 & indo1319 & & szl\_Latn & 112496 & indo1319 & & ksd\_Latn & 30550 & aust1307 & \\
tuk\_Latn & 649411 & turk1311 & & ish\_Latn & 111814 & atla1278 & & mzh\_Latn & 30517 & mata1289 & \\
tir\_Ethi & 649117 & afro1255 & & naq\_Latn & 109747 & khoe1240 & & llb\_Latn & 30480 & atla1278 & \\
tgk\_Latn & 636541 & indo1319 & & toh\_Latn & 107583 & atla1278 & & hra\_Latn & 30472 & sino1245 & \\
yua\_Latn & 610052 & maya1287 & & ttj\_Latn & 106925 & atla1278 & & mwm\_Latn & 30432 & cent2225 & \\
min\_Latn & 609065 & aust1307 & & nse\_Latn & 105189 & atla1278 & & krc\_Cyrl & 30353 & turk1311 & \\
lue\_Latn & 599429 & atla1278 & & hsb\_Latn & 104802 & indo1319 & & tuc\_Latn & 30349 & aust1307 & \\
khm\_Khmr & 590429 & aust1305 & yes & ami\_Latn & 104559 & aust1307 & & mrw\_Latn & 30304 & aust1307 & \\
tum\_Latn & 589857 & atla1278 & & alz\_Latn & 104392 & nilo1247 & & pls\_Latn & 30136 & otom1299 & \\
tll\_Latn & 586530 & atla1278 & & apc\_Arab & 102392 & afro1255 & & rap\_Latn & 30102 & aust1307 & \\
ekk\_Latn & 582595 & ural1272 & & vls\_Latn & 101900 & indo1319 & & fur\_Latn & 30052 & indo1319 & \\
lug\_Latn & 566948 & atla1278 & & mhr\_Cyrl & 100474 & ural1272 & & kaa\_Latn & 30031 & turk1311 & \\
niu\_Latn & 566715 & aust1307 & & djk\_Latn & 99234 & indo1319 & & prs\_Arab & 26823 & indo1319 & yes \\
tzo\_Latn & 540262 & maya1287 & & wes\_Latn & 98492 & indo1319 & & san\_Latn & 25742 & indo1319 & yes \\
mah\_Latn & 534614 & aust1307 & & gkn\_Latn & 97041 & atla1278 & & som\_Arab & 14199 & afro1255 & yes \\
tvl\_Latn & 521556 & aust1307 & & grc\_Grek & 96986 & indo1319 & & uig\_Latn & 9637 & turk1311 & yes \\
jav\_Latn & 516833 & aust1307 & yes & hbo\_Hebr & 96484 & afro1255 & & hau\_Arab & 9593 & afro1255 & yes \\
    \bottomrule
    \end{tabular}
}
    \caption{List of languages used to train \modelname{} (Part III).}
    \tablabel{languages3}
\end{table*}

\section{List of \dsources}
\label{sec:appendix_datasets}
The datasets and repositories used in this project involve: 
AI4Bharat,\footnote{\url{https://ai4bharat.org/}}
AIFORTHAI-LotusCorpus,\footnote{\url{https://github.com/korakot/corpus/releases/download/v1.0/AIFORTHAI-LotusCorpus.zip}}
Add \citep{el-haj-etal-2018-arabic},
AfriBERTa \citep{ogueji2021small},
AfroMAFT \citep{adelani-etal-2022-thousand,xue-etal-2021-mt5},
Anuvaad,\footnote{\url{https://github.com/project-anuvaad/anuvaad-parallel-corpus}}
AraBench \citep{sajjad-etal-2020-arabench},
AUTSHUMATO,\footnote{\url{https://autshumato.sourceforge.net/}}
Bloom \citep{DBLP:conf/emnlp/LeongNMFOW22}, 
CC100 \citep{conneau-etal-2020-unsupervised,DBLP:conf/lrec/WenzekLCCGJG20}, 
CCNet \citep{wenzek-etal-2020-ccnet}, 
CMU\_Haitian\_Creole,\footnote{\url{http://www.speech.cs.cmu.edu/haitian/text/}}
CORP.NCHLT,\footnote{\url{https://repo.sadilar.org/handle/20.500.12185/7}}
Clarin,\footnote{\url{https://www.clarin.si/}}
DART \citep{alsarsour-etal-2018-dart},
Earthlings \citep{DBLP:journals/lre/Dunn20}, 
FFR,\footnote{\url{https://github.com/bonaventuredossou/ffr-v1/tree/master/FFR-Dataset}}
Flores200 \citep{costa2022no}, 
GiossaMedia \citep{gongora-etal-2022-use, gongora-etal-2021-experiments},
Glosses \citep{camacho-collados-etal-2016-large},
Habibi \citep{el-haj-2020-habibi},
HinDialect \citep{bafna2022empirical}, 
HornMT,\footnote{\url{https://github.com/asmelashteka/HornMT}}
IITB \citep{kunchukuttan-etal-2018-iit},
IndicNLP \citep{nakazawa-etal-2021-overview},
Indiccorp \citep{kakwani-etal-2020-indicnlpsuite}, 
isiZulu,\footnote{\url{https://zenodo.org/record/5035171}}
JParaCrawl \citep{morishita-etal-2020-jparacrawl}, 
KinyaSMT,\footnote{\url{https://github.com/pniyongabo/kinyarwandaSMT}}
LeipzigData \citep{DBLP:conf/lrec/GoldhahnEQ12}, 
Lindat,\footnote{\url{https://lindat.cz/faq-repository}}
Lingala\_Song\_Lyrics,\footnote{\url{https://github.com/espoirMur/songs_lyrics_webscrap}}
Lyrics,\footnote{\url{https://lyricstranslate.com/}}
MC4 \citep{DBLP:journals/jmlr/RaffelSRLNMZLL20}, 
MTData \citep{gowda-etal-2021-many}, 
MaCoCu \citep{DBLP:conf/eamt/BanonEFGKLNSRRS22}, 
Makerere MT Corpus,\footnote{\url{https://zenodo.org/record/5089560}}
Masakhane community,\footnote{\url{https://github.com/masakhane-io/masakhane-community}}
Mburisano\_Covid,\footnote{\url{https://repo.sadilar.org/handle/20.500.12185/536}}
Menyo20K \citep{adelani-etal-2021-effect},
Minangkabau corpora \citep{koto-koto-2020-towards}, 
MoT \citep{palen-michel-etal-2022-multilingual},
NLLB\_seed \citep{costa2022no},
Nart/abkhaz,\footnote{\url{https://huggingface.co/datasets/Nart/abkhaz_text}}
OPUS \citep{TIEDEMANN12.463}, 
OSCAR \citep{suarez2019asynchronous}, 
ParaCrawl \citep{banon-etal-2020-paracrawl},
Parallel Corpora for Ethiopian Languages \citep{abate-etal-2018-parallel}, 
Phontron \citep{neubig11kftt},
QADI \citep{abdelali-etal-2021-qadi},
Quechua-IIC \citep{zevallos-etal-2022-introducing}, 
SLI\_GalWeb.1.0 \citep{agerri-etal-2018-developing},
Shami \citep{abu-kwaik-etal-2018-shami},
Stanford NLP,\footnote{\url{https://nlp.stanford.edu/}}
StatMT,\footnote{\url{https://statmt.org/}}
TICO \citep{anastasopoulos-etal-2020-tico},
TIL \citep{mirzakhalov-etal-2021-large},
Tatoeba,\footnote{\url{https://tatoeba.org/en/}}
TeDDi \citep{moran-etal-2022-teddi},
Tilde \citep{rozis-skadins-2017-tilde},
W2C \citep{11858/00-097C-0000-0022-6133-9}, 
WAT \citep{nakazawa-etal-2022-overview},
WikiMatrix \citep{schwenk-etal-2021-wikimatrix}, 
Wikipedia,\footnote{\url{https://huggingface.co/datasets/wikipedia}}
Workshop on NER for South and South East Asian Languages \citep{singh-2008-named},
XLSum \citep{DBLP:conf/acl/HasanBIMLKRS21}.

\section{Results for Each Task and Language}
\label{sec:appendix_results}
We report the detailed results for all tasks and languages in Table~\ref{tab:tatoeba} (Sentence Retrieval Tatoeba), \ref{tab:bible1}, \ref{tab:bible2} (Sentence Retrieval Bible), \ref{tab:ner} (NER), and \ref{tab:pos} (POS), \ref{tab:tc1}, \ref{tab:tc2} (Text Classification), \ref{tab:round-trip1}, \ref{tab:round-trip2} (Round Trip Alignment).

\begin{table*}
\centering
    \resizebox{\textwidth}{!}{
\begin{tabular}{cccc|cccc|cccc}
    \toprule
    \languagescriptcap &XLM-R-B &XLM-R-L &\modelname &\languagescriptcap &XLM-R-B &XLM-R-L &\modelname &\languagescriptcap &XLM-R-B &XLM-R-L &\modelname \\ \midrule
    afr\_Latn &71.9 &76.5 &\textbf{81.1} &heb\_Hebr &76.3 &\textbf{84.1} &76.0 &pam\_Latn &4.8 &5.6 &\textbf{11.0} \\
    amh\_Ethi &35.1 &37.5 &\textbf{44.6} &hin\_Deva &73.8 &\textbf{88.8} &85.6 &pes\_Arab &83.3 &86.6 &\textbf{87.6} \\
    ara\_Arab &59.2 &\textbf{66.8} &64.2 &hrv\_Latn &79.6 &85.6 &\textbf{89.8} &pms\_Latn &16.6 &12.6 &\textbf{54.5} \\
    arz\_Arab &32.5 &47.8 &\textbf{63.5} &hsb\_Latn &21.5 &23.0 &\textbf{53.6} &pol\_Latn &82.6 &\textbf{89.6} &82.4 \\
    ast\_Latn &59.8 &59.8 &\textbf{87.4} &hun\_Latn &76.1 &\textbf{81.8} &69.2 &por\_Latn &91.0 &\textbf{92.1} &90.1 \\
    aze\_Latn &62.6 &78.3 &\textbf{79.9} &hye\_Armn &64.6 &40.0 &\textbf{83.2} &ron\_Latn &86.0 &\textbf{89.1} &82.8 \\
    bel\_Cyrl &70.0 &80.5 &\textbf{81.4} &ido\_Latn &25.7 &28.8 &\textbf{57.6} &rus\_Cyrl &89.6 &\textbf{91.6} &91.5 \\
\bottomrule
    \end{tabular}
    }
    \caption{Top10 accuracy of XLM-R-B, XLM-R-L, and \modelname{} on Sentence Retrieval Tatoeba.}
    \tablabel{tatoeba}
\end{table*}

\begin{table*}
\centering
    \resizebox{\textwidth}{!}{
    \begin{tabular}{cccc|cccc|cccc}
        \toprule
    \languagescriptcap &XLM-R-B &XLM-R-L &\modelname &\languagescriptcap &XLM-R-B &XLM-R-L &\modelname &\languagescriptcap &XLM-R-B &XLM-R-L &\modelname \\ \midrule
ace\_Latn &4.4 &4.6 &\textbf{53.4} &iba\_Latn &14.4 &13.6 &\textbf{66.0} &pan\_Guru &43.2 &\textbf{59.4} &48.8 \\
ach\_Latn &4.4 &3.2 &\textbf{40.0} &ibo\_Latn &5.0 &3.0 &\textbf{30.4} &pap\_Latn &12.4 &9.2 &\textbf{72.4} \\
acr\_Latn &2.6 &3.4 &\textbf{25.4} &ifa\_Latn &4.4 &4.4 &\textbf{39.2} &pau\_Latn &4.4 &4.0 &\textbf{29.8} \\
afr\_Latn &76.8 &\textbf{77.2} &69.4 &ifb\_Latn &4.8 &3.6 &\textbf{36.6} &pcm\_Latn &13.6 &10.4 &\textbf{66.8} \\
agw\_Latn &5.8 &3.0 &\textbf{36.0} &ikk\_Latn &3.0 &3.2 &\textbf{50.6} &pdt\_Latn &9.2 &8.6 &\textbf{68.6} \\
ahk\_Latn &3.0 &2.6 &\textbf{3.2} &ilo\_Latn &6.2 &3.6 &\textbf{55.0} &pes\_Arab &69.4 &72.2 &\textbf{80.8} \\
aka\_Latn &5.0 &4.2 &\textbf{57.0} &ind\_Latn &\textbf{82.6} &80.4 &72.2 &pis\_Latn &6.4 &5.0 &\textbf{57.2} \\
\bottomrule
\end{tabular}
    }
    \caption{Top10 accuracy of XLM-R-B, XLM-R-L, and \modelname{} on Sentence Retrieval Bible  (Part I).}
    \tablabel{bible1}
\end{table*}

\begin{table*}
\centering
    \resizebox{\textwidth}{!}{
    \begin{tabular}{cccc|cccc|cccc}
        \toprule
    \languagescriptcap &XLM-R-B &XLM-R-L &\modelname &\languagescriptcap &XLM-R-B &XLM-R-L &\modelname &\languagescriptcap &XLM-R-B &XLM-R-L &\modelname \\ \midrule
dtp\_Latn &5.4 &4.2 &\textbf{24.2} &mdy\_Ethi &2.8 &2.4 &\textbf{31.6} &tih\_Latn &5.2 &4.4 &\textbf{51.6} \\
dyu\_Latn &4.2 &2.4 &\textbf{50.2} &meu\_Latn &5.6 &4.4 &\textbf{52.0} &tir\_Ethi &7.4 &6.2 &\textbf{43.4} \\
dzo\_Tibt &2.2 &2.0 &\textbf{36.4} &mfe\_Latn &9.0 &6.8 &\textbf{78.6} &tlh\_Latn &7.8 &6.4 &\textbf{72.4} \\
efi\_Latn &4.4 &4.2 &\textbf{54.0} &mgh\_Latn &5.2 &3.4 &\textbf{23.6} &tob\_Latn &2.2 &3.0 &\textbf{16.8} \\
ell\_Grek &52.6 &\textbf{53.8} &48.6 &mgr\_Latn &4.0 &4.4 &\textbf{57.6} &toh\_Latn &4.0 &4.0 &\textbf{47.2} \\
enm\_Latn &39.8 &39.2 &\textbf{66.0} &mhr\_Cyrl &6.6 &5.4 &\textbf{48.0} &toi\_Latn &4.2 &4.4 &\textbf{47.4} \\
epo\_Latn &\textbf{64.6} &59.8 &56.2 &min\_Latn &9.4 &6.2 &\textbf{29.0} &toj\_Latn &4.2 &4.0 &\textbf{15.6} \\
\bottomrule
\end{tabular}
    }
    \caption{Top10 accuracy of XLM-R-B, XLM-R-L, and \modelname{} on Sentence Retrieval Bible (Part II).}
    \tablabel{bible2}
\end{table*}

\begin{table*}
\centering
    \resizebox{\textwidth}{!}{

    \begin{tabular}{cccc|cccc|cccc}
      \toprule
      \languagescriptcap &XLM-R-B &XLM-R-L &\modelname &\languagescriptcap &XLM-R-B &XLM-R-L &\modelname &\languagescriptcap &XLM-R-B &XLM-R-L &\modelname \\ \midrule
        ace\_Latn &33.4 &38.9 &\textbf{44.2} &heb\_Hebr &51.5 &\textbf{56.5} &49.0 &ori\_Orya &\textbf{31.4} &27.6 &31.0 \\
afr\_Latn &75.6 &\textbf{78.3} &76.7 &hin\_Deva &67.0 &\textbf{71.1} &69.4 &oss\_Cyrl &33.7 &39.2 &\textbf{52.1} \\
als\_Latn &60.7 &61.4 &\textbf{80.0} &hrv\_Latn &77.2 &\textbf{78.9} &77.3 &pan\_Guru &50.0 &\textbf{50.5} &48.1 \\
amh\_Ethi &42.2 &40.9 &\textbf{45.4} &hsb\_Latn &64.0 &69.0 &\textbf{71.2} &pms\_Latn &71.2 &74.9 &\textbf{75.9} \\
ara\_Arab &44.7 &48.7 &\textbf{56.1} &hun\_Latn &76.2 &\textbf{79.8} &75.9 &pnb\_Arab &57.0 &64.6 &\textbf{65.8} \\
arg\_Latn &73.6 &74.6 &\textbf{77.2} &hye\_Armn &50.8 &\textbf{61.7} &54.8 &pol\_Latn &77.5 &\textbf{81.2} &78.1 \\
arz\_Arab &48.3 &52.5 &\textbf{57.4} &ibo\_Latn &40.8 &42.8 &\textbf{58.6} &por\_Latn &77.8 &\textbf{81.2} &78.6 \\
\bottomrule
\end{tabular}
    }
    \caption{F1 of XLM-R-B, XLM-R-L, and \modelname{} on NER.}
    \tablabel{ner}
\end{table*}

\begin{table*}
\centering
    \resizebox{\textwidth}{!}{

    \begin{tabular}{cccc|cccc|cccc}
      \toprule
      \languagescriptcap &XLM-R-B &XLM-R-L &\modelname &\languagescriptcap &XLM-R-B &XLM-R-L &\modelname &\languagescriptcap &XLM-R-B &XLM-R-L &\modelname \\ \midrule
        afr\_Latn &88.7 &\textbf{89.3} &87.5 &hbo\_Hebr &38.9 &45.7 &\textbf{54.2} &pol\_Latn &84.7 &\textbf{85.4} &82.4 \\
ajp\_Arab &62.9 &67.3 &\textbf{69.7} &heb\_Hebr &68.0 &\textbf{69.2} &67.2 &por\_Latn &88.6 &\textbf{89.8} &88.2 \\
aln\_Latn &53.5 &\textbf{60.4} &52.3 &hin\_Deva &71.3 &\textbf{75.3} &70.3 &quc\_Latn &28.9 &29.3 &\textbf{62.4} \\
amh\_Ethi &64.5 &\textbf{66.2} &66.1 &hrv\_Latn &85.9 &\textbf{86.2} &85.5 &ron\_Latn &83.9 &\textbf{85.7} &80.6 \\
ara\_Arab &68.5 &\textbf{69.7} &65.4 &hsb\_Latn &71.5 &74.4 &\textbf{83.6} &rus\_Cyrl &89.1 &\textbf{89.7} &88.7 \\
bam\_Latn &25.4 &23.5 &\textbf{40.8} &hun\_Latn &82.6 &\textbf{82.7} &81.2 &sah\_Cyrl &20.3 &22.8 &\textbf{76.8} \\
bel\_Cyrl &86.2 &\textbf{86.2} &86.0 &hye\_Armn &85.2 &\textbf{86.5} &84.0 &san\_Deva &18.3 &\textbf{28.6} &26.1 \\
\bottomrule
\end{tabular}
    }
    \caption{F1 of XLM-R-B, XLM-R-L, and \modelname{} on POS.}
    \tablabel{pos}
\end{table*}

\begin{table*}
\centering
    \resizebox{\textwidth}{!}{

    \begin{tabular}{cccc|cccc|cccc}
      \toprule
      \languagescriptcap &XLM-R-B &XLM-R-L &\modelname &\languagescriptcap &XLM-R-B &XLM-R-L &\modelname &\languagescriptcap &XLM-R-B &XLM-R-L &\modelname \\ \midrule
      ace\_Latn &15 &25 &\textbf{60} &iba\_Latn &30 &35 &\textbf{56} &ote\_Latn &6 &5 &\textbf{36} \\
ace\_Latn &15 &25 &\textbf{60} &iba\_Latn &30 &35 &\textbf{56} &ote\_Latn &6 &5 &\textbf{36} \\
ach\_Latn &9 &8 &\textbf{34} &ibo\_Latn &8 &6 &\textbf{51} &pag\_Latn &22 &21 &\textbf{52} \\
acr\_Latn &10 &8 &\textbf{46} &ifa\_Latn &12 &12 &\textbf{47} &pam\_Latn &20 &18 &\textbf{41} \\
afr\_Latn &54 &\textbf{64} &57 &ifb\_Latn &14 &11 &\textbf{48} &pan\_Guru &53 &\textbf{65} &59 \\
agw\_Latn &11 &13 &\textbf{54} &ikk\_Latn &11 &7 &\textbf{47} &pap\_Latn &31 &36 &\textbf{55} \\
ahk\_Latn &5 &5 &\textbf{24} &ilo\_Latn &15 &13 &\textbf{52} &pau\_Latn &12 &10 &\textbf{41} \\
\bottomrule
\end{tabular}
    }
    \caption{F1 of XLM-R-B, XLM-R-L, and \modelname{} on Text Classification  (Part I).}
    \tablabel{tc1}
\end{table*}

\begin{table*}
\centering
    \resizebox{\textwidth}{!}{

    \begin{tabular}{cccc|cccc|cccc}
      \toprule
      \languagescriptcap &XLM-R-B &XLM-R-L &\modelname &\languagescriptcap &XLM-R-B &XLM-R-L &\modelname &\languagescriptcap &XLM-R-B &XLM-R-L &\modelname \\ \midrule
efi\_Latn &10 &9 &\textbf{50} &meu\_Latn &15 &11 &\textbf{52} &tha\_Thai &56 &\textbf{67} &61 \\
ell\_Grek &37 &47 &\textbf{54} &mfe\_Latn &16 &14 &\textbf{61} &tih\_Latn &11 &11 &\textbf{56} \\
eng\_Latn &74 &\textbf{75} &68 &mgh\_Latn &10 &6 &\textbf{35} &tir\_Ethi &23 &27 &\textbf{48} \\
enm\_Latn &46 &56 &\textbf{65} &mgr\_Latn &14 &12 &\textbf{46} &tlh\_Latn &30 &26 &\textbf{59} \\
epo\_Latn &53 &\textbf{63} &53 &mhr\_Cyrl &14 &10 &\textbf{43} &tob\_Latn &6 &9 &\textbf{52} \\
est\_Latn &62 &\textbf{68} &53 &min\_Latn &27 &37 &\textbf{50} &toh\_Latn &11 &8 &\textbf{41} \\
eus\_Latn &28 &\textbf{33} &22 &miq\_Latn &7 &7 &\textbf{48} &toi\_Latn &14 &10 &\textbf{40} \\
\bottomrule
\end{tabular}
    }
    \caption{F1 of XLM-R-B, XLM-R-L, and \modelname{} on Text Classification  (Part II).}
    \tablabel{tc2}
\end{table*}

\begin{table*}
\centering
    \resizebox{\textwidth}{!}{
    \begin{tabular}{cccc|cccc|cccc}
      \toprule
      \languagescriptcap &XLM-R-B &XLM-R-L &\modelname &\languagescriptcap &XLM-R-B &XLM-R-L &\modelname &\languagescriptcap &XLM-R-B &XLM-R-L &\modelname \\ \midrule
ace\_Latn &2.50 &2.83 &\textbf{4.56} &hye\_Armn &2.32 &3.25 &\textbf{4.91} &pam\_Latn &2.85 &3.52 &\textbf{4.46} \\
ach\_Latn &3.13 &4.02 &\textbf{5.60} &hye\_Latn &2.34 &\textbf{2.98} &2.44 &pan\_Guru &2.11 &2.73 &\textbf{4.11} \\
acr\_Latn &2.01 &2.46 &\textbf{2.51} &iba\_Latn &2.77 &3.85 &\textbf{6.01} &pap\_Latn &3.12 &3.85 &\textbf{5.46} \\
afr\_Latn &3.17 &3.66 &\textbf{5.46} &ibo\_Latn &2.05 &2.43 &\textbf{4.33} &pau\_Latn &2.67 &3.09 &\textbf{4.09} \\
agw\_Latn &2.51 &2.80 &\textbf{4.09} &ifa\_Latn &1.81 &2.40 &\textbf{3.45} &pcm\_Latn &3.81 &4.44 &\textbf{6.47} \\
ahk\_Latn &1.11 &\textbf{1.23} &1.22 &ifb\_Latn &2.22 &2.58 &\textbf{3.28} &pdt\_Latn &2.41 &3.33 &\textbf{5.11} \\
aka\_Latn &3.38 &4.50 &\textbf{6.48} &ikk\_Latn &1.75 &2.29 &\textbf{3.83} &pes\_Arab &2.66 &3.91 &\textbf{4.81} \\
\bottomrule
\end{tabular}
    }
    \caption{Accuracy of XLM-R-B, XLM-R-L, and \modelname{} on Round Trip Alignment (Part I). }
    \tablabel{round-trip1}
\end{table*}

\begin{table*}
\centering
    \resizebox{\textwidth}{!}{
    \begin{tabular}{cccc|cccc|cccc}
      \toprule
      \languagescriptcap &XLM-R-B &XLM-R-L &\modelname &\languagescriptcap &XLM-R-B &XLM-R-L &\modelname &\languagescriptcap &XLM-R-B &XLM-R-L &\modelname \\ \midrule
efi\_Latn &2.55 &3.25 &\textbf{6.23} &mfe\_Latn &3.61 &4.19 &\textbf{6.26} &toi\_Latn &3.19 &4.10 &\textbf{4.31} \\
ell\_Grek &2.79 &3.38 &\textbf{4.77} &mgh\_Latn &2.78 &3.28 &\textbf{3.48} &toj\_Latn &1.43 &1.84 &\textbf{2.25} \\
eng\_Latn &4.02 &4.49 &\textbf{6.39} &mgr\_Latn &3.32 &4.06 &\textbf{6.39} &ton\_Latn &2.01 &2.64 &\textbf{3.63} \\
enm\_Latn &3.77 &4.60 &\textbf{7.19} &mhr\_Cyrl &2.75 &3.28 &\textbf{5.32} &top\_Latn &1.56 &2.16 &\textbf{2.19} \\
epo\_Latn &4.01 &4.83 &\textbf{5.88} &min\_Latn &2.62 &3.05 &\textbf{3.78} &tpi\_Latn &2.44 &2.71 &\textbf{5.96} \\
est\_Latn &4.34 &5.24 &\textbf{8.21} &miq\_Latn &2.23 &3.13 &\textbf{4.12} &tpm\_Latn &2.79 &3.39 &\textbf{4.67} \\
eus\_Latn &3.12 &3.80 &\textbf{4.19} &mkd\_Cyrl &3.99 &4.54 &\textbf{7.37} &tsn\_Latn &2.82 &3.12 &\textbf{4.63} \\
\bottomrule
\end{tabular}
    }
    \caption{Accuracy of XLM-R-B, XLM-R-L, and \modelname{} on Round Trip Alignment  (Part II).}
    \tablabel{round-trip2}
\end{table*}

\section{Perplexity Results for all Languages}
\label{sec:appendix_perplexity_results}

Perplexity number for all languages is
presented in Table~\ref{tab:perplexities1}, Table~\ref{tab:perplexities2},
and Table~\ref{tab:perplexities3}.

\begin{table*}
    \centering
    \resizebox{\textwidth}{!}{
        \begin{tabular}{cccc|cccc|cccc}
        \toprule
        Language-Script & XLM-R-B & XLM-R-L & \modelname & Language-Script & XLM-R-B & XLM-R-L & \modelname & Language-Script & XLM-R-B & XLM-R-L & \modelname \\
        \midrule
        srd\_Latn & 87.2 & 66.6 & \textbf{5.4} & aka\_Latn & 86.7 & 74.1 & \textbf{14.2} & dyu\_Latn & 68.5 & 27.4 & \textbf{10.2} \\
ben\_Beng & 5.2 & \textbf{3.7} & 7.2 & mon\_Latn & 288 & 282.4 & \textbf{33.7} & nyy\_Latn & 628.5 & 198.3 & \textbf{18.0} \\
ajp\_Arab & 74.6 & \textbf{34.0} & 44.8 & gor\_Latn & 89.8 & 140.7 & \textbf{8.8} & tzh\_Latn & 320.3 & 82.8 & \textbf{4.7} \\
tdx\_Latn & 688.4 & 716.4 & \textbf{16.0} & kjb\_Latn & 110.8 & 81.1 & \textbf{16.2} & hne\_Deva & 80.1 & 60.3 & \textbf{9.1} \\
tpm\_Latn & 99.9 & 90.2 & \textbf{17.9} & lhu\_Latn & 44.7 & 12.3 & \textbf{2.0} & bel\_Cyrl & 3.4 & \textbf{2.5} & 5.3 \\
grc\_Grek & 10.1 & 10.4 & \textbf{3.4} & bos\_Latn & 6.1 & \textbf{3.4} & 7.9 & szl\_Latn & 46.4 & 30.2 & \textbf{3.1} \\
sxn\_Latn & 469.2 & 148.3 & \textbf{14.5} & lmo\_Latn & 48.4 & 25.9 & \textbf{6.1} & ksh\_Latn & 340.3 & 227.6 & \textbf{19.9} \\
\bottomrule
    \end{tabular}
}
    \caption{Perplexity of all languages covered by \modelname{} (Part I).}
    \tablabel{perplexities1}
\end{table*}

\begin{table*}
    \centering
    \resizebox{\textwidth}{!}{
        \begin{tabular}{cccc|cccc|cccc}
        \toprule
        Language-Script & XLM-R-B & XLM-R-L & \modelname & Language-Script & XLM-R-B & XLM-R-L & \modelname & Language-Script & XLM-R-B & XLM-R-L & \modelname \\
        \midrule

        swc\_Latn & 39.2 & 22.5 & \textbf{13.2} & top\_Latn & 589.2 & 89.6 & \textbf{23.5} & hin\_Latn & \textbf{11.1} & 22.1 & 11.9 \\
deu\_Latn & 4.4 & \textbf{3.6} & 10.2 & bin\_Latn & 278.1 & 169.8 & \textbf{13.3} & eng\_Latn & 5.7 & \textbf{4.0} & 7.5 \\
caq\_Latn & 185.9 & 129 & \textbf{21.6} & chw\_Latn & 778.9 & 645.8 & \textbf{33.9} & hus\_Latn & 134.6 & 68.2 & \textbf{5.3} \\
ceb\_Latn & 63.1 & 53.1 & \textbf{2.1} & hyw\_Cyrl & 268.5 & 233.5 & \textbf{6.3} & urh\_Latn & 236.8 & 211.5 & \textbf{11.4} \\
nia\_Latn & 280.3 & 85.5 & \textbf{7.5} & kor\_Hang & 7.2 & \textbf{2.6} & 11 & mkd\_Cyrl & 4.3 & \textbf{3.1} & 6.2 \\
urd\_Arab & 8.3 & \textbf{5.3} & 8.7 & btx\_Latn & 463 & 163.1 & \textbf{19.3} & wbm\_Latn & 58.9 & 47.3 & \textbf{13.6} \\
niu\_Latn & 600.1 & 437.5 & \textbf{10.1} & srn\_Latn & 609.3 & 137.2 & \textbf{12.6} & kwn\_Latn & 1053.6 & 753.2 & \textbf{32.0} \\
\bottomrule
    \end{tabular}
}
    \caption{Perplexity of all languages covered by \modelname{} (Part II).}
    \tablabel{perplexities2}
\end{table*}

\begin{table*}
    \centering
    \resizebox{\textwidth}{!}{
        \begin{tabular}{cccc|cccc|cccc}
        \toprule
        Language-Script & XLM-R-B & XLM-R-L & \modelname & Language-Script & XLM-R-B & XLM-R-L & \modelname & Language-Script & XLM-R-B & XLM-R-L & \modelname \\
        \midrule

        bts\_Latn & 205.7 & 204.5 & \textbf{8.8} & tsn\_Latn & 264.7 & 137.8 & \textbf{12.5} & orm\_Latn & 23.4 & \textbf{8.6} & 16 \\
gla\_Latn & 11.5 & 12.7 & \textbf{7.2} & pon\_Latn & 928.4 & 181.9 & \textbf{19.2} & luo\_Latn & 699.4 & 258.5 & \textbf{85.1} \\
kat\_Latn & 36.4 & 24.8 & \textbf{18.3} & nmf\_Latn & 297.6 & 310.6 & \textbf{44.9} & pcm\_Latn & 38.3 & 169.6 & \textbf{3.6} \\
uig\_Latn & 188.8 & 173.9 & \textbf{15.2} & ajg\_Latn & 147.1 & 149.5 & \textbf{22.6} & nnb\_Latn & 364.1 & 95 & \textbf{28.6} \\
kat\_Geor & 6 & \textbf{3.9} & 6.4 & tir\_Ethi & 28.3 & 15.7 & \textbf{4.4} & kaz\_Cyrl & \textbf{4.3} & 5.4 & 9.6 \\
mlg\_Latn & 10.9 & \textbf{4.4} & 7.6 & bhw\_Latn & 411.2 & 126.2 & \textbf{21.6} & dzo\_Tibt & 8.5 & \textbf{3.3} & 5.7 \\
arn\_Latn & 382.7 & 96.7 & \textbf{17.6} & mhr\_Cyrl & 122.9 & 168.4 & \textbf{5.8} & sun\_Latn & 23.6 & \textbf{11.9} & 17 \\
\bottomrule
    \end{tabular}
}
    \caption{Perplexity of all languages covered by \modelname{} (Part III).}
    \tablabel{perplexities3}
\end{table*}

\end{document}